\documentclass[journal]{IEEEtran}

\usepackage{amsmath,amssymb}
\usepackage{graphicx}
\usepackage{booktabs}
\usepackage{makecell}
\usepackage{multirow}
\usepackage{multicol}
\usepackage{cite}
\usepackage{url}
\usepackage[table]{xcolor}
\usepackage{algorithmic}
\usepackage{algorithm}
\usepackage{array}

\begin{document}
\title{SinD 2.0: A Multi-City UAV Dataset with Semantic Risk Annotations for SOTIF-Oriented Safety Validation at Signalized Intersections}

\author{Yunwei~Li,
Shengjie~Fu,
Chunrong~Chen,
Chengxiang~Zhao,
Yuchen~Fan,
Mingyu~Zhu,
Yanchao~Xu,
Jiahui Xu,
Anran Wang,
Huanan Wang,
Yuxin~Zhang,
Lan~Yang,
Chuzhao~Li,
Jie~Ji,
Yi~He,
Abhijit~Sarkar,
Akash~Sonth,
Hong~Wang, and
Jun~Li%
\thanks{Yunwei Li, Shengjie Fu, Chunrong Chen, Chengxiang Zhao, Mingyu Zhu, Huanan Wang, Hong Wang, and Jun Li are with Tsinghua University, Beijing, China (e-mail: li-yw23@mails.tsinghua.edu.cn; zhao\_cx25@mails.tsinghua.edu.cn; zhumy23@mails.tsinghua.edu.cn; hong\_wang@tsinghua.edu.cn; lijun1958@tsinghua.edu.cn).}%
\thanks{Yuchen Fan is with Beijing Institute of Technology, Beijing, China (e-mail: 1120233198@bit.edu.cn).}%
\thanks{Yanchao Xu is with Guangzhou Automobile Group Co., Ltd., Guangzhou, China (e-mail: xuyanchao@gac.com.cn).}%
\thanks{Jiahui Xu is with the Department of Data and Systems Engineering, the University of Hong Kong, China (e-mail: jiahui.xu@connect.hku.hk).}%
\thanks{Anran Wang is with the School of Information Science and Engineering, East China University of Science and Technology, Shanghai, China (e-mail: anran19931072017@163.com).}%
\thanks{Yuxin Zhang is with Jilin University, Changchun, China (e-mail: yuxinzhang@jlu.edu.cn).}%
\thanks{Lan Yang is with Chang'an University, Xi'an, China (e-mail: lanyang@chd.edu.cn).}%
\thanks{Chuzhao Li is with Chongqing University, Chongqing, China (e-mail: lichuzhao@cqu.edu.cn).}%
\thanks{Jie Ji is with Southwest University, Chongqing, China (e-mail: jijiess@swu.edu.cn).}%
\thanks{Yi He is with Wuhan University of Technology, Wuhan, China (e-mail: heyi@whut.edu.cn).}%
\thanks{Abhijit Sarkar is with Virginia Tech Transportation Institute, Blacksburg, VA, USA (e-mail: asarkar@vtti.vt.edu).}%
\thanks{Akash Sonth is with Virginia Tech, Blacksburg, VA, USA (e-mail: akashsonth@vt.edu).}}

\markboth{SinD 2.0 Draft Manuscript}%
{Li: SinD 2.0}
\maketitle

\begin{abstract}
Safety validation at signalized intersections remains a critical bottleneck for the deployment of autonomous driving systems (ADS), as these scenarios involve dense heterogeneous traffic, contested right of way, and long-tail safety-critical interactions, posing significant challenges to the Safety of the Intended Functionality (SOTIF). Existing naturalistic driving datasets often suffer from geographical homogeneity, sparsity of safety-critical events, and lack of semantic risk annotations, which limit the evaluation of algorithmic generalizability and targeted SOTIF verification. To address these gaps, this paper introduces SinD 2.0, a large-scale drone-based intersection dataset dedicated to cross-domain ADS safety analysis.

The main contributions of SinD 2.0 are: (1) Cross-domain diversity: It covers six signalized intersections across four Chinese cities, capturing distinct intersection topologies and regional driving behavior characteristics; (2) High-density risk interactions: A total of 32,682 safety-critical events are extracted via surrogate safety measures, significantly enriching the density of boundary test scenarios; (3) Hierarchical semantic annotations: Besides integration with high-definition (HD) maps and Signal Phase and Timing (SPaT) data, it provides multi-dimensional semantic labels including traffic violations, high-risk interactions, visual shielding, and narrow feasible areas; (4) Full-stack testing toolchain: It supports automated scenario extraction, prediction-only evaluation, open-loop replay, reactive closed-loop testing, and photorealistic rendering.
Benchmark experiments demonstrate that SinD 2.0 exhibits significant domain shifts across cities, and the semantic risk subsets can effectively expose the performance limitations of ADS algorithms. The dataset, annotations, and testing toolchain are available at \url{https://github.com/SOTIF-AVLab/SinD/tree/main}.
\end{abstract}


\IEEEpeerreviewmaketitle

\begin{figure*}[t]
    \centering
    \includegraphics[width=0.93\textwidth]{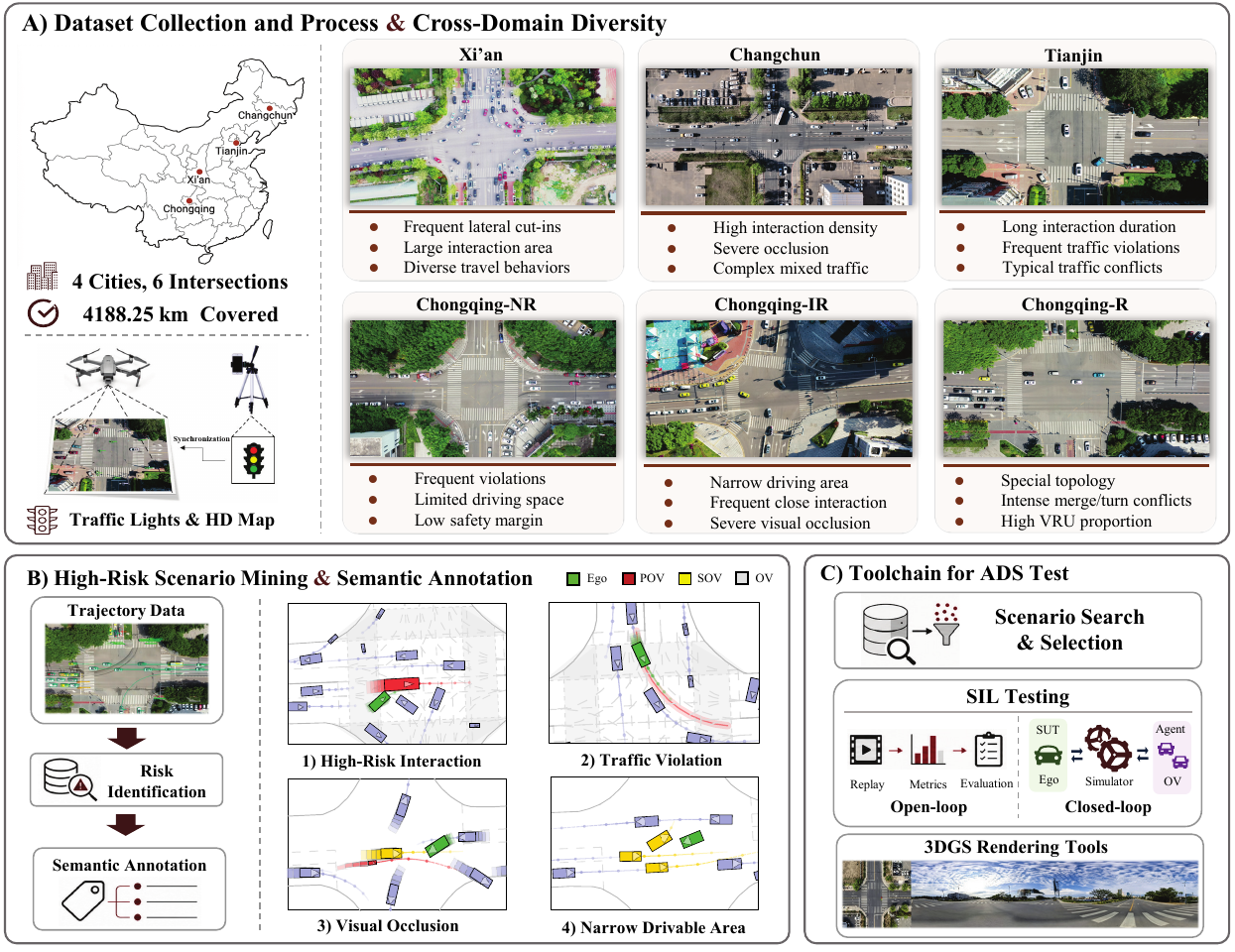} 
    \caption{Overview of the SinD 2.0 dataset construction, semantic annotation, and toolchain. (A) Geographic distribution of the six intersections across four cities, showcasing cross-domain diversity and intersection-specific risk tags. (B) The pipeline for high-risk scenario mining and multi-level semantic annotation including visual occlusion and narrow drivable area. (C) The comprehensive toolchain enabling scenario search, open-loop and closed-loop SIL testing, and rendering.}
    \label{fig:overall_teaser}
\end{figure*}

\section{Introduction}
As Autonomous Driving Systems (ADS) advance, navigating complex urban environments, particularly signalized intersections, has emerged as a primary operational bottleneck. These intersections feature high-density heterogeneous traffic, spatiotemporal right-of-way contentions, and unpredictable multi-agent interactions. Consequently, addressing the Safety of the Intended Functionality (SOTIF) in dynamic environments is a major challenge for the widespread adoption of ADS.

The core methodology for addressing SOTIF challenges relies on scenario-based testing and verification. Within this paradigm, real-world traffic datasets serve as a foundational resource. They provide critical references for ensuring the naturalistic properties of simulated driving behaviors and achieving logical coverage of operational scenarios. These datasets also supply the data needed to develop and train ADS algorithms, such as data-driven trajectory prediction, planning, and scenario generation models. Ultimately, the quality and diversity of the dataset directly determine the limits of ADS testing and development.

However, from the perspective of safety validation, existing trajectory and aerial datasets exhibit limitations that hinder their application. A primary conflict exists between the demand for comprehensive scenario coverage and the geographical homogeneity of current datasets. Rigorous testing requires scenario diversity and extensive Operational Design Domain (ODD) coverage to ensure algorithmic robustness. By contrast, current datasets are frequently constrained to a single city or a few regional sites. This geographical homogeneity fails to capture the cross-domain heterogeneity of driving cultures, compliance rates, and road topologies, limiting the ability to evaluate ADS generalization across different urban domains.

Effective safety validation also requires a high degree of criticality, which conflicts with the sparsity of dangerous scenarios in naturalistic driving data. Rigorous evaluation must push the system to its performance boundaries, yet naturalistic datasets are dominated by safe, routine interactions. High-value, safety-critical scenarios, including near-misses, intense conflicts, and aggressive maneuvers, are exceedingly rare. This long-tail distribution renders current datasets inefficient for boundary testing, risk evaluation, and the training of generative corner-case models.

Evaluating SOTIF also requires exposing the system to scenarios induced by specific semantic triggering conditions, a feature lacking in existing datasets. SOTIF verification often depends on complex triggers such as severe visual occlusions, narrow feasible spaces, or sudden intent changes by Vulnerable Road Users (VRUs). Most current datasets provide only raw spatiotemporal trajectories and basic bounding boxes. They lack the multi-level semantic annotations needed to identify, retrieve, and reconstruct specific causal events. Without explicit labels connecting kinematic data to risk factors, deep safety analysis and targeted scenario extraction remain restricted.

To overcome these barriers, we introduce SinD 2.0, a large-scale, drone-based traffic dataset designed for cross-domain ADS safety analysis. Covering six complex signalized intersections across four cities in China, SinD 2.0 captures the kinematic states of heterogeneous traffic from a bird's-eye view, featuring high-density interactions and cross-domain characteristics. To facilitate SOTIF-oriented analysis and verification, the dataset includes hierarchical semantic annotations for multiple safety-critical events. We also provide a toolchain for scenario extraction, simulation, scenario rendering, and metric computation to support ADS testing (Fig.~\ref{fig:overall_teaser}).

The main contributions are summarized as follows:
\begin{itemize}
\item \textbf{Cross-Domain Diversity and Driving Style Heterogeneity}: By spanning six intersections in four geographically distinct cities, SinD 2.0 captures a wide spectrum of intersection topologies and quantifiable regional driving behaviors. This variability establishes a foundation for evaluating ADS behavior under varied operating conditions and studying generalization across different domains.
\item \textbf{High Density and Variety of Safety-Critical Scenarios}: Addressing the sparsity of naturalistic data, SinD 2.0 features dense critical corner cases, including aggressive traffic violations and intense VRU conflicts. This concentration of high-risk events directly supports SOTIF testing and the rigorous evaluation of safety-critical responses.
\item \textbf{Multiple Semantic Annotations for Interaction Events}: The dataset integrates HD Maps, Signal Phase and Timing (SPaT), and hierarchical semantic annotations for safety-critical events alongside raw trajectories. These labels connect kinematic interactions to specific SOTIF triggering conditions, facilitating targeted scenario retrieval and safety analysis.
\item \textbf{Comprehensive Toolchain for Testing and Evaluation}: To streamline safety validation, we provide a toolchain supporting automated scenario selection, open-loop and closed-loop evaluations, and an auxiliary rendering pipeline for vision-oriented testing. This infrastructure enables researchers to integrate the dataset into SOTIF verification workflows and systematically inspect policy behavior.
\end{itemize}

The remainder of this paper is organized as follows. Section~\ref{sec:relatedwork} reviews the related literature. Section~\ref{sec:basic_info} introduces the construction and fundamental characteristics of the dataset. Section~\ref{sec:analysis} analyzes the multi-intersection characteristics of the dataset. Section~\ref{sec:semantic} details the extraction and annotation of specific semantic risk scenarios. Section~\ref{sec:experiments} presents the accompanying toolchain and simulation experiments for autonomous driving safety validation. Section~\ref{sec:conclusion} concludes the paper.

\section{Related Work}\label{sec:relatedwork}

\subsection{Naturalistic Trajectory Datasets for Autonomous Driving}
High-quality naturalistic driving datasets are the bedrock of autonomous driving research. Representative ego-centric datasets, such as KITTI \cite{geiger2012we}, nuScenes \cite{caesar2020nuscenes}, and Waymo Open Dataset \cite{sun2020scalability}, primarily rely on ego-vehicle sensor suites such as cameras and LiDARs. Argoverse further advanced map-based forecasting by coupling trajectory annotations with rich semantic maps \cite{chang2019argoverse}. The Waymo Open Motion Dataset extended this direction to large-scale interactive motion forecasting \cite{ettinger2021large}. While these datasets have significantly advanced perception and onboard prediction algorithms, they are fundamentally constrained by the ego-vehicle's limited field-of-view and pervasive occlusions. Consequently, they struggle to capture the complete global topology and multi-agent interaction dynamics necessary for intersection analysis.

To overcome these limitations, bird's-eye view (BEV) trajectory datasets collected via drones or infrastructure cameras have gained prominence. Pioneering datasets like NGSIM \cite{punzo2011assessment} and highD \cite{krajewski2018highd} provided massive trajectory data but focused predominantly on structured highway segments, lacking the complex cross-traffic and unprotected maneuvers in urban driving conditions. Datasets such as inD \cite{bock2020ind} and the INTERACTION dataset \cite{zhan2019interaction} shifted the focus to intersections and roundabouts, offering valuable insights into urban interactions. rounD provides high-resolution drone-based trajectories at German roundabouts \cite{krajewski2020round}, while openDD further expands large-scale trajectory resources for roundabout scenarios \cite{breuer2020opendd}. CitySim emphasizes safety-oriented trajectory collection for digital-twin research \cite{zheng2022citysim}. Infrastructure-side datasets such as A9 also demonstrate the value of roadside sensing for capturing urban interactions beyond the ego-vehicle perspective \cite{zimmer2023a9}. Our previous work, SinD v1 \cite{xu2022sind}, also contributed to this domain by capturing high-fidelity trajectories at a Chinese signalized intersection. However, a critical gap remains: many existing datasets are still geographically concentrated, and their coverage of driving cultures, mixed-traffic composition, signal-control logic, and diverse topological constraints remains limited. This lack of cross-domain diversity severely limits their utility as benchmarks for evaluating the generalization and domain-adaptation capabilities of ADS algorithms.

\subsection{Safety-Critical Scenario Mining and SOTIF Analysis}
Scenario-based testing is widely recognized as an effective paradigm for validating ADS safety, particularly for addressing the Safety of the Intended Functionality (SOTIF) \cite{iso21448}. Recent studies further emphasize that scenario-based testing provides a practical bridge between naturalistic data, simulation, and safety validation \cite{zhong2021survey}. A core prerequisite for this testing is the extraction of safety-critical events (SCEs) from massive naturalistic data. Traditionally, researchers have relied on Surrogate Safety Measures (SSMs) such as Time-to-Collision (TTC), Post-Encroachment Time (PET), and Deceleration Rate to Avoid Crash (DRAC) to quantify interaction risks \cite{mahmud2017application}. These indicators are effective for screening potential conflicts, but they are limited in representing triggering mechanisms and behavioral intent.

Despite the maturity of these metrics, mining actionable scenarios from existing datasets still encounters two key challenges. First, naturalistic driving data exhibits a severe long-tail distribution; the vast majority of recorded trajectories represent safe, compliant driving, rendering high-value critical interactions extremely sparse~\cite{liu2024curse}. Accelerated evaluation studies have shown that targeted sampling is essential to expose rare but safety-relevant events \cite{zhao2016accelerated,ariefDeepProbabilisticAccelerated2021a}. Adaptive stress testing similarly indicates that systematic search is important for discovering uncommon failure cases \cite{koren2019adaptive}. Second, multi-level semantic annotations are currently absent from mainstream datasets, while such annotations are a core foundation for in-depth SOTIF assessment. Beyond simply identifying traffic conflicts, SOTIF validation aims to interpret their underlying causes, including visual occlusions, intense right-of-way disputes, and unexpected intrusions from vulnerable road users (VRU) \cite{zhao2020safety}. Surveys on safety-critical scenario generation also demonstrate that integrating trajectory-based conflict cues and high-level scenario semantics can effectively improve risk discovery performance \cite{ding2022survey}. With clear semantic labels for various triggering conditions, researchers are able to classify edge cases in a standardized manner and conduct causal safety analysis, which strongly empowers data-driven SOTIF verification work.

\subsection{Data-Driven Scenario Generation and Simulation Evaluation}
The emergence of generative AI and reinforcement learning has revolutionized ADS testing, transitioning it from static replay to reactive, closed-loop simulation \cite{feng2021testing}. Learning-based simulation methods such as SimNet model reactive traffic behavior from real-world observations \cite{bergamini2021simnet}, while BITS uses bi-level imitation to capture interactive driving behavior for traffic simulation \cite{xu2022bits}. TrafficGen focuses on generating diverse and realistic traffic scenarios for autonomous driving evaluation \cite{feng2022trafficgen}. Scenario Diffusion further explores controllable driving scenario generation through diffusion models \cite{pronovost2023scenario}. Recent studies in \emph{Automotive Innovation} have also shown complementary directions: one constructs signalized-intersection test scenarios from SinD through a traffic-participant model, and another uses large language models to translate natural-language requirements into executable autonomous-driving test scenarios \cite{gu2025generation,cai2026text2scenario}. Datasets equipped with abundant semantic descriptions and risk annotations serve as a solid foundation for these generative approaches. They support the synthesis of realistic corner cases and assist testing algorithms in locating critical performance limits.
Moreover, dedicated toolchains for converting raw trajectory data into simulation-ready scenario formats can substantially facilitate closed-loop evaluation. Microscopic traffic simulators such as SUMO have long been used to model network-level traffic dynamics and support large-scale traffic experiments \cite{krajzewicz2012recent}. Open urban driving simulators such as CARLA provide sensor-level rendering, vehicle dynamics, and interactive agents for closed-loop ADS testing \cite{dosovitskiy2017carla}. Scenario description standards and languages, including OpenSCENARIO and Scenic, further improve the reproducibility and portability of test cases across simulation platforms \cite{fremont2019scenic}. Benchmarking frameworks such as CommonRoad illustrate the importance of standardized scenario representations for motion planning evaluation \cite{althoff2017commonroad}. Recent platforms such as nuPlan and ScenarioNet further emphasize large-scale closed-loop evaluation built upon real-world driving scenarios \cite{karnchanachari2024nuplan,li2023scenarionet}. Nevertheless, preparing naturalistic trajectory data for these environments still requires nontrivial preprocessing, semantic mapping, and format conversion. By combining a risk-intensive dataset with a complete evaluation toolchain, SinD 2.0 directly addresses these practical demands.

\section{SinD 2.0 Dataset Construction and Fundamental Characteristics}\label{sec:basic_info}

This section details the construction process and statistical characteristics of the SinD 2.0 dataset. To address the geographical and cultural homogeneity of many existing intersection datasets, we established a drone-based data collection and processing pipeline that covers multiple cities and intersection topologies. 
Following this, we present statistical analyses and dataset comparisons that characterize the coverage and risk density of SinD 2.0.

\subsection{Data Collection and Scenario Distribution}

To capture cross-domain heterogeneity and multi-agent interactions in Chinese urban traffic, SinD 2.0 includes six signalized intersections across four cities. These cities span varied terrains, climates, and regional driving cultures, providing a testbed for studying the generalization of autonomous driving algorithms. 

As shown in Fig.~\ref{fig:overall_teaser}(A), the environmental characteristics and risk patterns differ across these sites. Changchun presents high interaction density and visual occlusions from large vehicles, creating a complex mixed-traffic environment. Tianjin features a large grid intersection with prolonged interactions and frequent traffic violations, making it suitable for extracting typical conflicts. Xi'an has a highly mixed traffic flow where frequent lateral encroachment by VRUs is associated with large interaction areas and high behavioral variance. Chongqing includes three intersections with distinct risk modalities under complex terrain. The first involves narrow drivable areas, close-proximity interactions, and visual occlusions. The second features compressed spaces associated with low-margin evasive maneuvers. The third has a non-orthogonal topology with turning and merging conflicts and a high volume of VRUs. 

For data collection, we used hovering high-altitude drones to record 4K top-down video. This bird's-eye perspective avoids the visual blind spots and dynamic occlusions of onboard sensors, capturing global intersection topologies and the complete kinematic states of all traffic participants.

% TODO，给External vehicle-view SCE Ratio 也加上 $\sim$
\begin{table*}[t]
\centering
\caption{Comparison of SinD 2.0 with mainstream autonomous driving datasets. }
\resizebox{\textwidth}{!}{%
\begin{tabular}{lllcrcrrcccc}
\toprule
Dataset & Platform & Location / Scenarios & Tracks & \makecell{Duration \\(h)} & \makecell{VRU \\(\%)} & TL Info & Map & Risk Labels & SCE Count & SCE Ratio & \makecell{SCE Density\\(Events/h)} \\ \midrule
NuScenes~\cite{caesar2020nuscenes} & Vehicle & Boston, SG / Urban & $\sim$90k & 320 & 20.1 & Yes & Yes & No & 1,486 & 49.29\% & 313.31 \\
Waymo~\cite{ettinger2021large} & Vehicle & USA / Urban & 7.6M & 574 & 11.5 & Yes & Yes & No & $\sim$145k & 40.64\% & 1319.68 \\
Argoverse 2~\cite{wilson2023argoverse2} & Vehicle & USA / Urban & 13.9M & 763 & 10.0 & Yes & Yes & No & $\sim$115k & 48.66\% & 1506.11 \\
NuPlan~\cite{karnchanachari2024nuplan} & Vehicle & USA, SG / Urban & $\sim$5M & 1282 & 46.3 & Yes & Yes & No & $\sim$43k & 18.61\% & 101.15 \\ \midrule
HighD~\cite{krajewski2018highd} & Drone & Germany / Highway & 110k & 16.5 & 0 & No & No & No & 827 & 6.92\% & 49.65 \\
InD~\cite{bock2020ind} & Drone & Germany / Intersection & 11.5k & 10 & $\sim$43.0 & No & Yes & No & 1,573 & 16.58\% & 160.20 \\
INTERACTION~\cite{zhan2019interaction} & Drone & Global / Intersection & 40k & 16.5 & $\sim$5.0 & Yes & Yes & No & 19,478 & 75.09\% & 1524.48 \\
SinD~\cite{xu2022sind} & Drone & Tianjin, CN / Intersection & 13.2k & 7 & 62.6 & Yes & Yes & No & 7,790 & 48.68\% & 1120.39 \\ \midrule
\rowcolor{gray!15} SinD 2.0 (Ours) & Drone & 4 Cities, CN / Intersection & 53k & 22.8 & 34.8 & Yes & Yes & Yes & 32,682 & 39.84\% & 1452.53 \\ \bottomrule
\multicolumn{12}{l}{External vehicle-view SCE metrics are approximate.} \\
\end{tabular}%
}
\label{tab:dataset_comparison}
\end{table*}

\subsection{Multi-Source Data Alignment}

Beyond high-fidelity trajectories, autonomous driving safety validation heavily relies on environmental context. Inheriting and upgrading the processing pipeline of SinD 1.0~\cite{xu2022sind}, SinD 2.0 aligns trajectories, maps, and signal states in a common spatiotemporal frame. The process begins with trajectory extraction and smoothing, where computer vision object detection and multi-object tracking algorithms extract the pixel coordinates of all dynamic traffic participants, including motor vehicles, non-motorized vehicles, and pedestrians, from the drone videos. These coordinates are projected into the world coordinate system using a calibration matrix, followed by recursive state filtering for trajectory smoothing and the calculation of kinematic parameters such as velocity and acceleration. Subsequently, high-definition maps are constructed by vectorizing static intersection infrastructure, incorporating topological and semantic attributes such as lane centerlines, boundaries, stop lines, and crosswalks to provide the foundation for subsequent violation determination and intent recognition. Finally, the signal phase and timing states are synchronized with the trajectory timestamps, which is crucial for reproducing intersection right-of-way logic and long-term trajectory prediction.

\subsection{Fundamental Statistics and Comparison}

Table~\ref{tab:dataset_comparison} presents a comparison with mainstream naturalistic driving datasets. Beyond fundamental dataset attributes, we examine the distribution of safety-critical events (SCE) from a SOTIF testing perspective. An interaction between a valid trajectory pair $(i, j)$ is classified as a safety-critical event based on the indicator function
\begin{equation}
\mathbb{I}_{\text{SCE}}(i, j) = \begin{cases} 1, & \text{if } \min\left(\text{PET}(i, j), \text{TTC}_{\text{robust}}(i, j)\right) < \tau \\ 0, & \text{otherwise} \end{cases}
\end{equation}
where the critical safety threshold $\tau$ is set to $3.0\,\text{s}$. To ensure the validity of the extracted events, we apply a filtering mechanism that automatically excludes stationary vehicles, non-interacting parallel driving, routine car-following behaviors, and measurement noise caused by low relative speeds. The specific mathematical definitions of these filtering thresholds and boundary conditions are detailed in Appendix~\ref{sec:appendix_sce}.

% The comparison should be interpreted with care: for external vehicle-view datasets, safety-critical event counts are estimated from available public splits and are therefore not a substitute for a fully uniform reprocessing of all raw data. 
We use the same high-level SCE definition and duration normalization where possible, but differences in sensor viewpoint, agent coverage, and released annotations make the comparison approximate. Under this protocol, SinD 2.0 provides about 53,000 tracked traffic participants across 22.8 hours of continuous recording, and its multi-city geographical span exceeds existing single-region drone datasets. While datasets such as INTERACTION report high safety-critical event ratios from selected bottleneck periods, SinD 2.0 uses continuous naturalistic recordings to capture traffic evolution from off-peak to congested states. This process extracts 32,682 high-risk interaction events and pairs trajectory-level criticality with semantic-level risk labels. Besides, SinD 2.0 adds SOTIF-oriented triggering conditions such as visual blind spots, rule deviations, and narrow feasible spaces through an automated annotation pipeline to facilitate comprehensive safety testing of ADS.

\begin{table}[h]
\centering
\caption{Total Distance Traveled by Modality in SinD 2.0}
\label{tab:distance_traveled}
\begin{tabular}{lrrr}
\toprule
Intersection & Total Dist. (km) & Vehicle (km) & VRU (km) \\ \midrule
Changchun (cc) & 836.90 & 828.82 & 8.08 \\
Tianjin (tj) & 667.39 & 481.27 & 186.12 \\
Chongqing-IR (cqIR) & 319.31 & 265.66 & 53.65 \\
Chongqing-NR (cqNR) & 208.31 & 173.58 & 34.73 \\
Chongqing-R (cqR) & 960.31 & 917.02 & 43.29 \\
Xi'an (xa) & 1196.04 & 1118.33 & 77.72 \\ \midrule
Overall Total & 4188.25 & 3784.67 & 403.58 \\ \bottomrule
\end{tabular}
\end{table}

As shown in Table~\ref{tab:distance_traveled}, SinD 2.0 encompasses a cumulative total of 4,188.25 kilometers of naturalistic trajectories. Notably, VRU trajectories—a defining feature of Chinese urban traffic—account for 403.58 kilometers. This extensive collection of high-density VRU data offers distinctive interaction scenarios absent from existing public benchmarks, providing valuable references for intent prediction and obstacle-avoidance planning in complex mixed-traffic environments.

% === Traffic participant type distributions ===
\begin{figure}[t]
    \centering
    \includegraphics[width=\columnwidth]{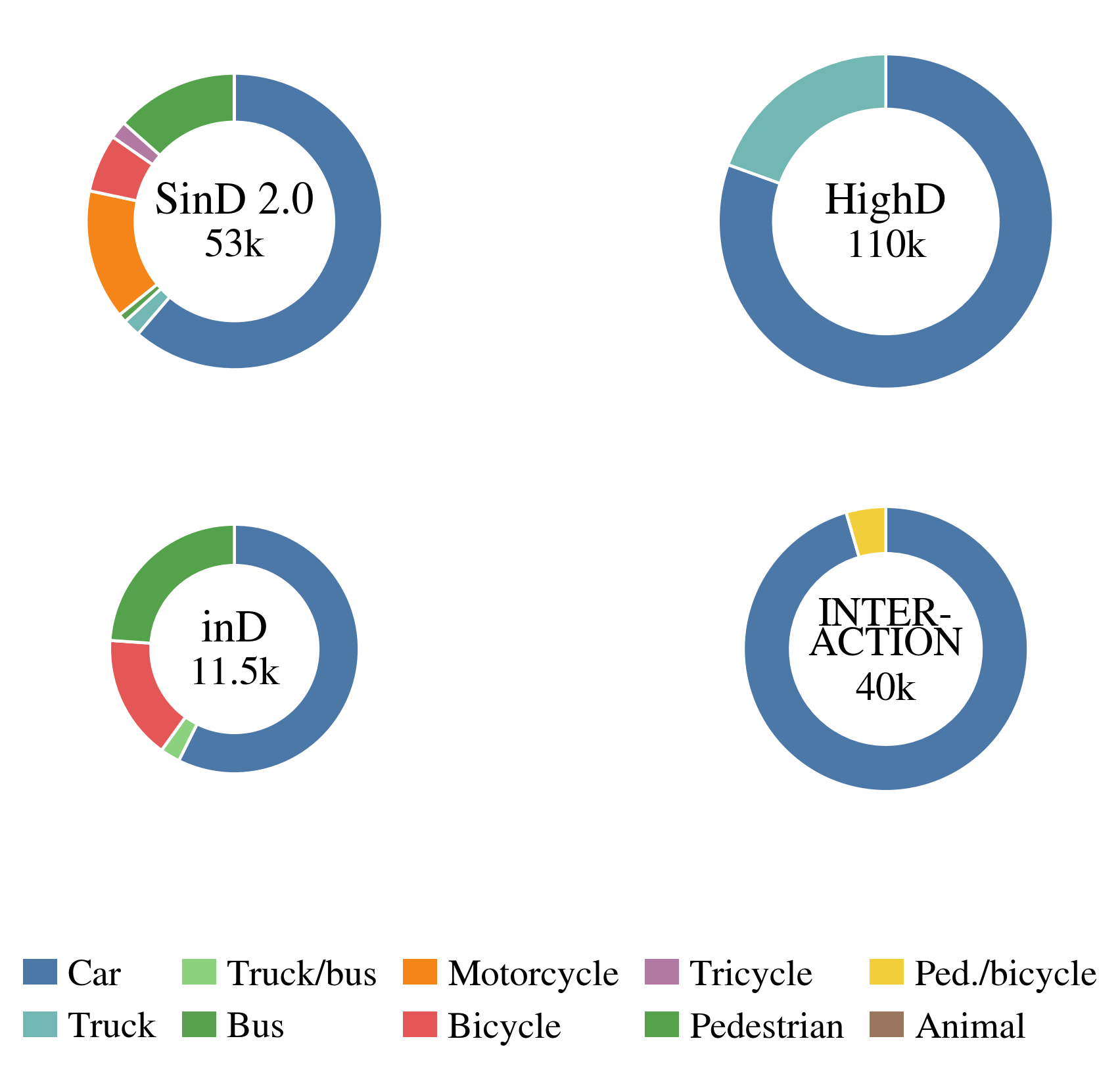}
    \caption{Comparison of traffic participant type distributions across SinD 2.0, highD, inD, and INTERACTION. The SinD 2.0 sector proportions are computed from the local class-level metadata.}
    \label{fig:tp_donut_comparison}
\end{figure}

% === SinD intersection participant type distributions ===
\begin{figure}[t]
    \centering
    \includegraphics[width=0.87\columnwidth]{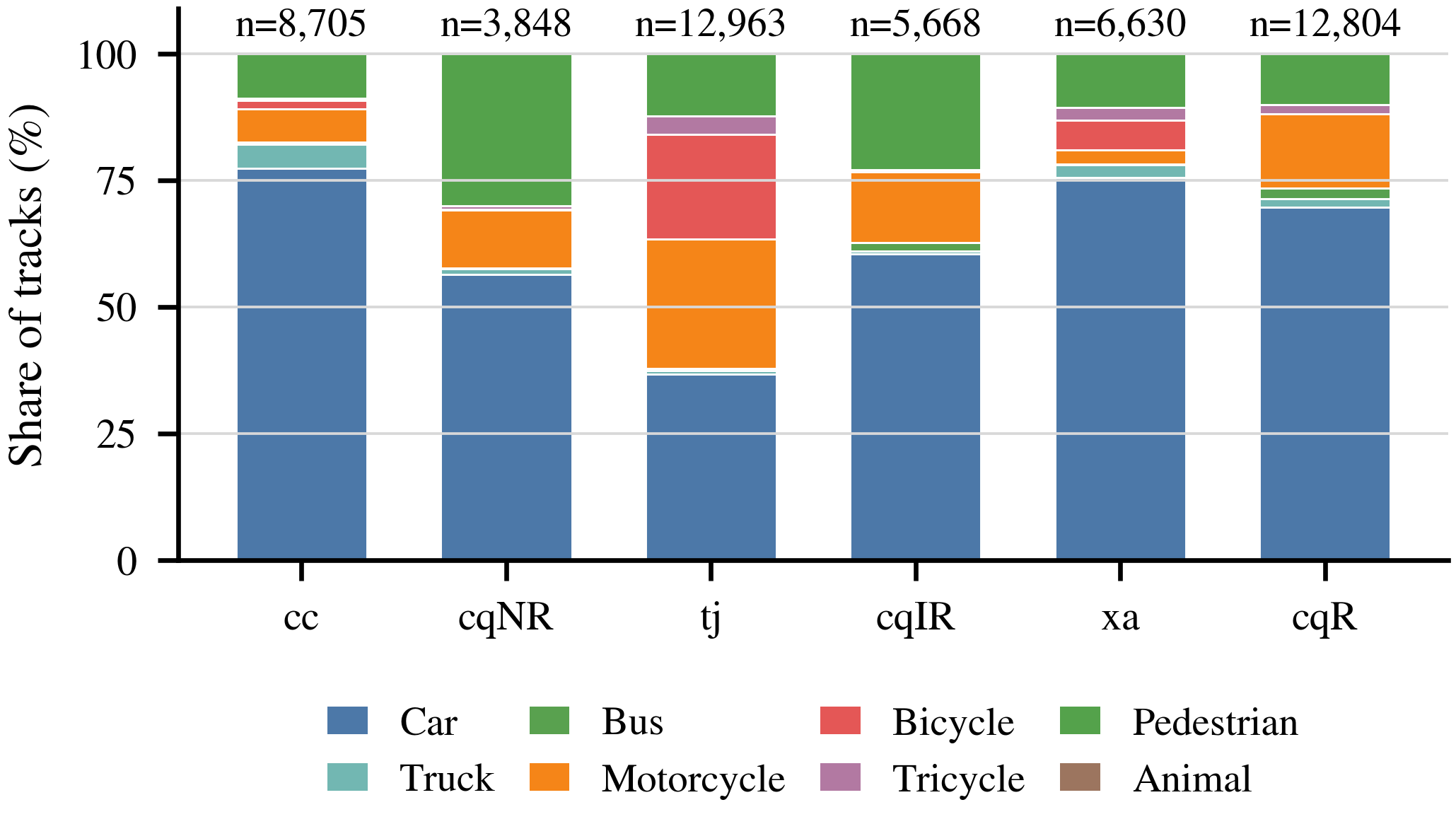}
    \caption{Traffic participant type composition across the six SinD 2.0 intersections. Each stacked bar reports the percentage of local participant trajectories by class, with the corresponding track count annotated above each bar.}
    \label{fig:sind_intersection_tp_distribution}
\end{figure}

\subsection{Heterogeneity and Composition across Intersections}

While Figure~\ref{fig:tp_donut_comparison} highlights the macroscopic diversity of SinD 2.0 compared to other benchmarks, Figure~\ref{fig:sind_intersection_tp_distribution} details how traffic participant modalities vary across individual collection sites. This distribution reveals that the dataset's heterogeneity is spatial and unevenly distributed, mirroring the distinct local risk profiles and driving environments of each intersection. 

Specifically, Tianjin emerges as the most two-wheeler-heavy site, characterized by large volumes of motorcycles and bicycles, whereas Chongqing-R and Changchun are predominantly defined by high vehicle densities. Conversely, Chongqing-IR and Chongqing-NR contribute high concentrations of pedestrian trajectories within highly compressed mountainous road geometries. Meanwhile, the Xi'an site captures extended vulnerable road user (VRU) crossings characterized by high lateral variance. Rather than presenting a uniform data distribution, SinD 2.0 provides a cross-domain testing paradigm where safety validation frameworks can expose a single navigation or prediction algorithm to varied operational mixtures ranging from vehicle-dominated flows to environments heavily influenced by two-wheelers or pedestrians.

\section{Cross-Domain Characteristics and Heterogeneity Analysis}\label{sec:analysis}

This section further analyzes why the six intersections should be regarded as distinct operational domains rather than as repeated samples of the same traffic scene. We organize the analysis around three complementary levels: static intersection topology, dynamic physical behavior, and interaction-level game complexity.

\subsection{Static Intersection Topology and Exposure}

\subsubsection{Geometric descriptors}
We first examine the relationship between static geometric properties and dynamic exposure. For each intersection, we define a core region of interest (ROI) around the un-channelized conflict area and compute the conflict area $A_c$, the core ROI area, and the angular skewness
\begin{equation}
\Delta_{\theta} = |90^{\circ} - \theta|,
\end{equation}
where $\theta$ denotes the principal intersection angle. We then correlate these geometric descriptors with dynamic exposure metrics, including ROI residence time, interaction degree, long-game ratio, and conflict-type composition.
Because the dataset contains six intersections, these correlations should be interpreted as descriptive evidence rather than as population-level significance tests. Nevertheless, the trends reveal how geometric design changes the type of safety problem faced by an ADS.

\begin{table}[t]
\centering
\caption{Definitions of Static Exposure Metrics Used in the Correlation Analysis}
\label{tab:static_exposure_metric_definitions}
\resizebox{\columnwidth}{!}{%
\begin{tabular}{lll}
\toprule
Metric & Definition & Unit \\ \midrule
Conflict area & Un-channelized intersection area where paths can overlap & m$^2$ \\
Core ROI area & Active central region used for residence and occupancy analysis & m$^2$ \\
Angle skewness & Absolute deviation of the principal intersection angle from $90^\circ$ & degree \\
ROI residence time & Time spent by a vehicle inside the core ROI & s \\
Mean interaction degree & Mean number of agents in an interaction component & agents \\
Long-game ratio & Share of interaction episodes lasting at least 5 s & \% \\
P90 interaction duration & 90th percentile of interaction episode duration & s \\
Crossing-conflict ratio & Share of classified pairwise conflicts that are crossing conflicts & \% \\
\bottomrule
\end{tabular}%
}
\end{table}

\begin{table}[t]
\centering
\caption{Static Geometric Descriptors of the Six Intersections}
\label{tab:static_geometry_descriptors}
\resizebox{\columnwidth}{!}{%
\begin{tabular}{lrrrr}
\toprule
Intersection & Conflict area (m$^2$) & Core ROI (m$^2$) & Angle ($^\circ$) & Skew ($^\circ$) \\ \midrule
Changchun (cc) & 691.91 & 633.93 & 90.0 & 0.0 \\
Tianjin (tj) & 1028.82 & 640.25 & 90.0 & 0.0 \\
Chongqing-IR (cqIR) & 1387.25 & 720.26 & 77.5 & 12.5 \\
Chongqing-NR (cqNR) & 1099.43 & 573.09 & 91.4 & 1.4 \\
Chongqing-R (cqR) & 2422.35 & 1052.75 & 86.8 & 3.2 \\
Xi'an (xa) & 2593.37 & 1381.11 & 86.2 & 3.8 \\
\bottomrule
\end{tabular}%
}
\end{table}

\begin{figure}[t]
    \centering
    \includegraphics[width=0.9\columnwidth]{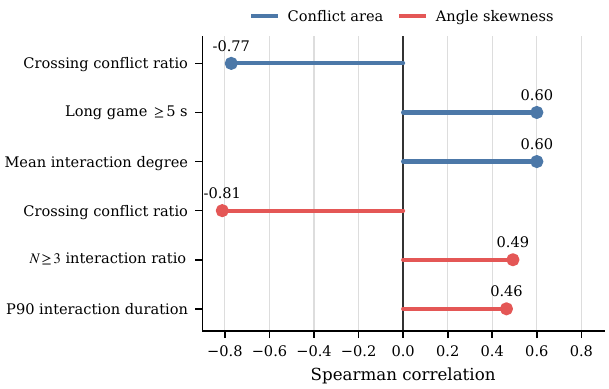}
    \caption{Core descriptive correlations between static geometric descriptors and exposure metrics across the six intersections. The plot retains only high-signal relationships with $|r|\geq0.5$.}
    \label{fig:geometric_correlation}
\end{figure}

\subsubsection{Un-channelized area and residence exposure}
Table~\ref{tab:static_geometry_descriptors} shows the static geometric descriptors for each intersection. Xi'an and Chongqing-R have the largest un-channelized conflict areas, while Chongqing-NR has the smallest core ROI. Figure~\ref{fig:geometric_correlation} further shows that larger conflict areas are associated with longer games and higher mean interaction degree, but with a lower crossing-conflict ratio. Thus, increasing open intersection space does not only create more perpendicular conflicts; it often converts a short crossing-risk problem into a longer planning problem in which agents have more possible yielding locations and more opportunities to enter multi-agent negotiations.

\begin{figure}[t]
    \centering
    \includegraphics[width=0.9\columnwidth]{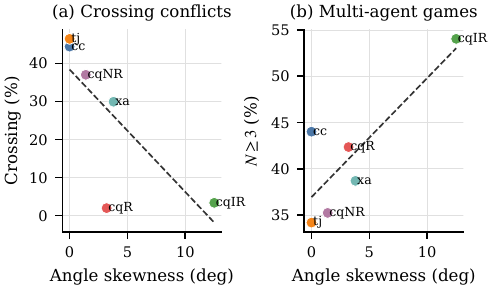}
    \caption{Effect of intersection angle skewness on crossing-conflict ratio and $N\geq3$ interaction ratio.}
    \label{fig:angle_skew_exposure}
\end{figure}

\subsubsection{Skewness as a negotiation amplifier}
Angular skewness provides a second geometric driver. As shown in Figure~\ref{fig:angle_skew_exposure}, skewness suppresses direct crossing-conflict dominance while increasing the share of multi-agent games. Chongqing-IR is the clearest example: its 12.5$^\circ$ skew produces the highest $N\geq3$ interaction ratio among the six intersections. Its P90 interaction duration is also the longest, indicating that skewed geometry is associated with extended negotiation episodes. Such scenarios are particularly challenging for forecasting models trained on regular grid-like intersections because ordinary right-of-way expectations, turning radii, and visibility relationships are distorted at the same time.

\subsubsection{Infrastructure-Induced VRU Exposure}

The physical infrastructure of the six intersections provides another layer of dataset diversity. All six sites contain unprotected left-turn movements, which are among the most important interaction patterns for SOTIF-oriented testing because they require vehicles to negotiate with opposing through traffic and crossing VRUs without a fully protected phase. At the same time, the VRU infrastructure is intentionally not uniform across the dataset. Several sites lack physical separation elements such as pedestrian refuge islands, lane dividers, or dedicated waiting areas inside the intersection. This absence does not simply change map appearance; it changes the feasible set of VRU trajectories and increases the chance that pedestrians or two-wheelers spill into vehicle conflict zones.

\begin{figure}[t]
    \centering
    \includegraphics[width=0.72\columnwidth]{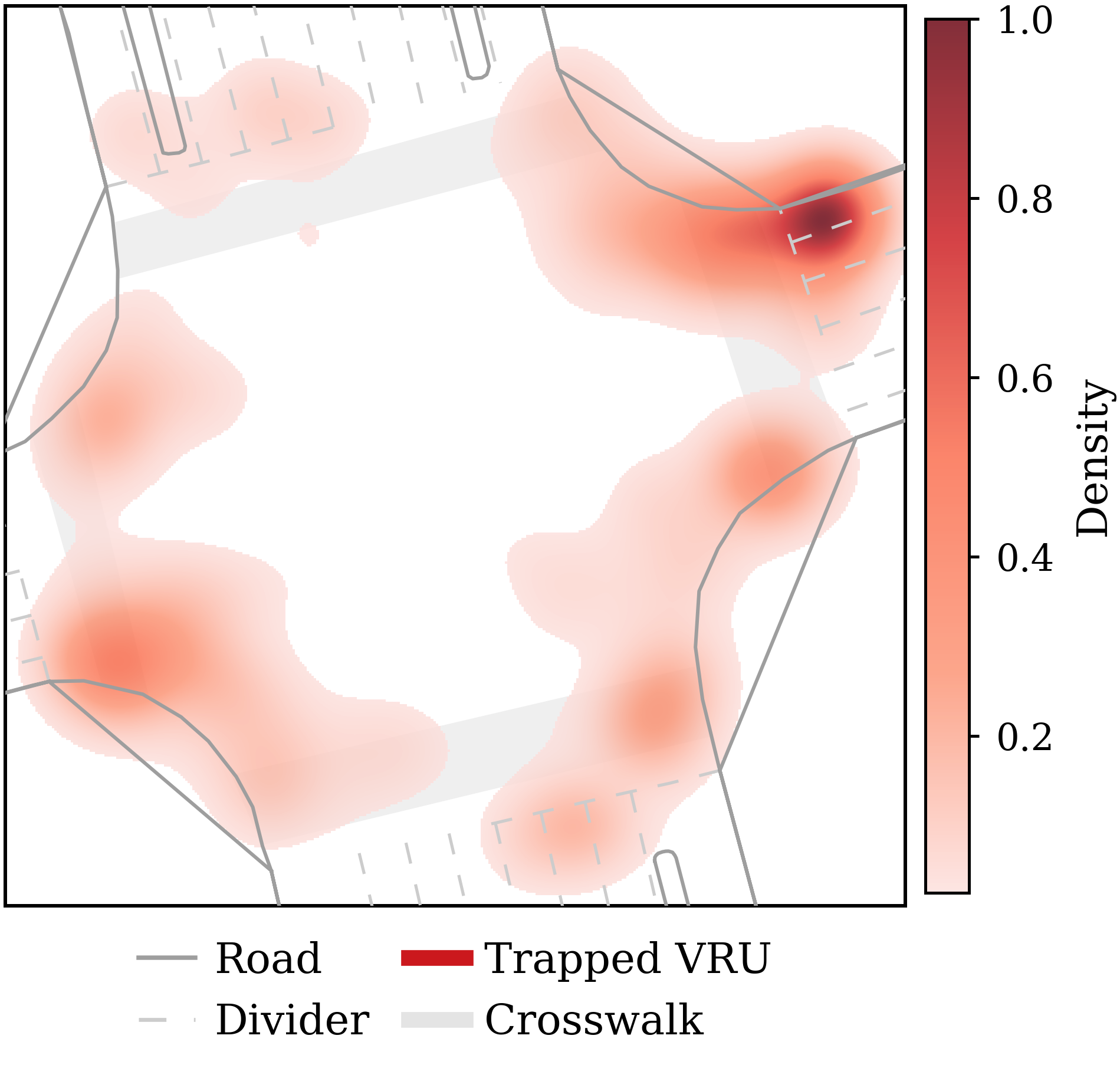}
    \caption{Example of infrastructure-induced VRU exposure at the Xi'an intersection. Long crossing distance and the lack of a clear central refuge area create a broad exposure region where slow or waiting VRUs can remain inside the motor-vehicle conflict zone.}
    \label{fig:xasl_trapped_vru}
\end{figure}

Figure~\ref{fig:xasl_trapped_vru} illustrates this phenomenon using the Xi'an site. The longest observed VRU crossing distance reaches 35.2 m, while no explicit central refuge island is available in the annotated map. Among 1,447 VRU trajectories analyzed at this site, the median displacement is 44.58 m, 939 trajectories travel at least 30 m, and 1,362 trajectories enter the central intersection region. Moreover, 266 trajectories exhibit low-speed residence of at least 3 s in the center region. These statistics are consistent with an infrastructure-related exposure pattern: limited physical separation is associated with VRU trajectories diffusing into the intersection core, long crossing distances prolong exposure, and signal changes can leave slow dynamic obstacles inside the conflict region. Such cases are difficult for ADS testing because they are neither simple static obstacles nor fully rule-following crossing agents. They represent dynamic, infrastructure-induced exposure events that require long-horizon prediction, cautious yielding, and robust deadlock handling.

\subsection{Dynamic Physical Characteristics}
We next characterize the dynamic physical properties of the six domains through four complementary indicators: (i) speed-acceleration envelopes, which describe the feasible kinematic boundary; (ii) lateral deviation from lane center lines, which measures unstructured turning behavior; (iii) accepted gaps in unprotected maneuvers, which reflect local aggressiveness; and (iv) spatiotemporal occupancy density, which reveals where vehicles accumulate and form congestion hot spots.

\begin{table}[t]
\centering
\caption{Summary of Dynamic Physical Characteristics}
\label{tab:dynamic_physical_summary}
\resizebox{\columnwidth}{!}{%
\begin{tabular}{lrrrr}
\toprule
Intersection & Env. area & Lat. var. (m$^2$) & Median gap (s) & ROI sec./veh. \\ \midrule
Changchun (cc) & 51.49 & 40.47 & 3.60 & 3.84 \\
Tianjin (tj) & 21.69 & 4.50 & 5.10 & 4.03 \\
Chongqing-IR (cqIR) & 27.16 & 4.10 & 6.40 & 3.48 \\
Chongqing-NR (cqNR) & 37.77 & 21.90 & 5.00 & 3.22 \\
Chongqing-R (cqR) & 36.89 & 3.60 & 7.55 & 3.29 \\
Xi'an (xa) & 40.75 & 12.86 & 4.60 & 5.89 \\
\bottomrule
\end{tabular}%
}
\end{table}

Table~\ref{tab:dynamic_physical_summary} summarizes the four physical indicators used in this subsection, which confirms that the six sites expose different motion limits, lane-adherence patterns, maneuver aggressiveness, and accumulation regions.

\subsubsection{Kinematic envelope}
The kinematic envelope measures the site-level diversity of observed speed--acceleration states. A larger envelope indicates that vehicles operate across a broader range of motion regimes, including higher-speed passages and stronger acceleration or braking, whereas a compact envelope suggests more concentrated and conservative behavior. We compute this envelope from all valid trajectory points at each intersection. For each site, a 95\% kernel-density envelope is estimated in the $(v,a)$ plane, and cross-site overlap is used to quantify whether two domains induce similar kinematic regimes.

\begin{figure}[t]
    \centering
    \includegraphics[width=0.9\columnwidth]{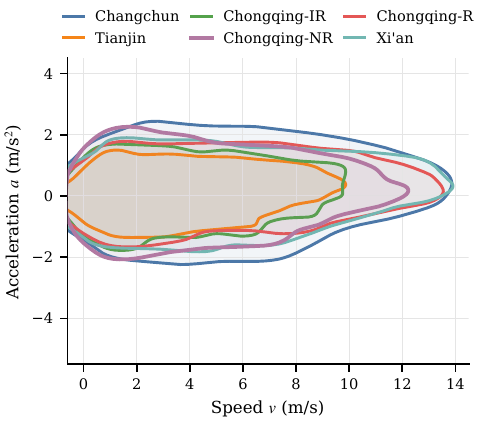}
    \caption{Cross-city speed-acceleration envelopes estimated from trajectory points.}
    \label{fig:kinematic_envelopes}
\end{figure}

Figure~\ref{fig:kinematic_envelopes} illustrates the kinematic envelopes of the six intersections and shows substantial cross-domain differences in motion-state distributions. Changchun has the largest envelope area, indicating the greatest diversity of vehicle motion and the presence of more aggressive or high-speed driving states. Chongqing-R, Chongqing-NR, and Xi'an also cover relatively high speed ranges, although their acceleration spread is more concentrated than Changchun. Tianjin and Chongqing-IR are comparatively compact, suggesting more concentrated and conservative motion regimes.

\subsubsection{Lateral Deviation from Reference Lanes}
Lateral deviation quantifies the degree to which turning vehicles adhere to the reference lane geometry. Greater deviation variance indicates weaker lane-center priors and more unstructured turning behavior, which implies an elevated potential for traffic conflicts. This metric is computed by matching candidate turning trajectories to reference lanes and measuring their lateral offsets along the arc length of the reference line. Among 13,518 candidate turning trajectories, 6,523 can be matched to reference lanes within a 6 m threshold, forming 204 distinct turning clusters. Open or weakly channelized turning regions produce broader lateral deviation distributions, whereas more constrained sites confine most turning maneuvers closer to the reference lane. For intent prediction, such behavioral patterns degrade the reliability of lane-centerline priors; for motion planning, they widen the uncertainty tube around the nominal path. Collectively, these observations yield a rich set of test cases for ADS safety testing and facilitate systematic test scenario generation.

\begin{figure}[t]
    \centering
    \includegraphics[width=0.9\columnwidth]{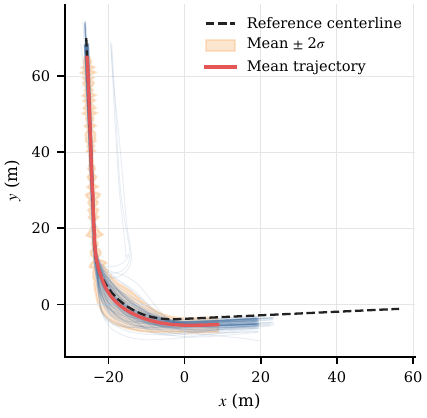}
    \caption{Representative lateral-deviation bundle for a Changchun left-turn movement. Raw trajectories are shown with low opacity, together with the reference centerline, a reference-line-reconstructed mean trajectory, and a mean $\pm2\sigma$ lateral spread band.}
    \label{fig:lateral_deviation}
\end{figure}

Figure~\ref{fig:lateral_deviation} illustrates this effect for a Changchun left-turn movement. The broad band around the mean shows that the same semantic maneuver can occupy multiple lateral modes inside the intersection core, which is the relevant stressor for lane-conditioned forecasting of ADS.

\subsubsection{Accepted gaps in unprotected maneuvers}
Accepted gap measures the temporal margin that drivers are willing to accept in unprotected conflict passages. Smaller accepted gaps indicate more aggressive yielding behavior and provide a risk-relevant boundary condition for testing prediction and planning models. We mine accepted gaps by detecting cases in which a turning vehicle passes through a conflict point before an opposing vehicle arrives. The lower tail is the most safety-relevant part of the distribution because it defines realistic aggressive-yielding boundaries that rule-based traffic models often fail to reproduce.

\begin{figure}[t]
    \centering
    \includegraphics[width=0.9\columnwidth]{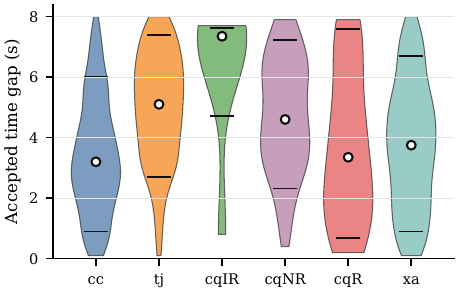}
    \caption{Accepted time-gap distributions for unprotected conflict passages. The violin shape shows the distributional spread, with median and 10--90\% markers overlaid.}
    \label{fig:critical_gap}
\end{figure}

Figure~\ref{fig:critical_gap} compares the accepted-gap distributions across locations. Changchun and Xi'an contain many low-gap accepted maneuvers, whereas the Chongqing-IR and Chongqing-R samples are sparse and shifted toward larger gaps. This suggests that gap acceptance is entangled with topology, visibility, and available maneuver space rather than being a universal driver parameter.

\subsubsection{Spatiotemporal occupancy density}
Spatiotemporal occupancy density identifies where exposure accumulates inside the intersection core. High-density cells indicate locations where vehicles repeatedly slow, queue, or become locally locked, while a diffuse pattern indicates that exposure is spread across a larger maneuvering area. We compute this metric by rasterizing the core ROI into 1 m $\times$ 1 m cells and accumulating the total vehicle occupancy time in each cell. It complements the scalar ROI residence values in Table~\ref{tab:dynamic_physical_summary} by showing whether exposure is spatially concentrated or diffused across the intersection core.

\begin{figure}[t]
    \centering
    \includegraphics[width=0.9\columnwidth]{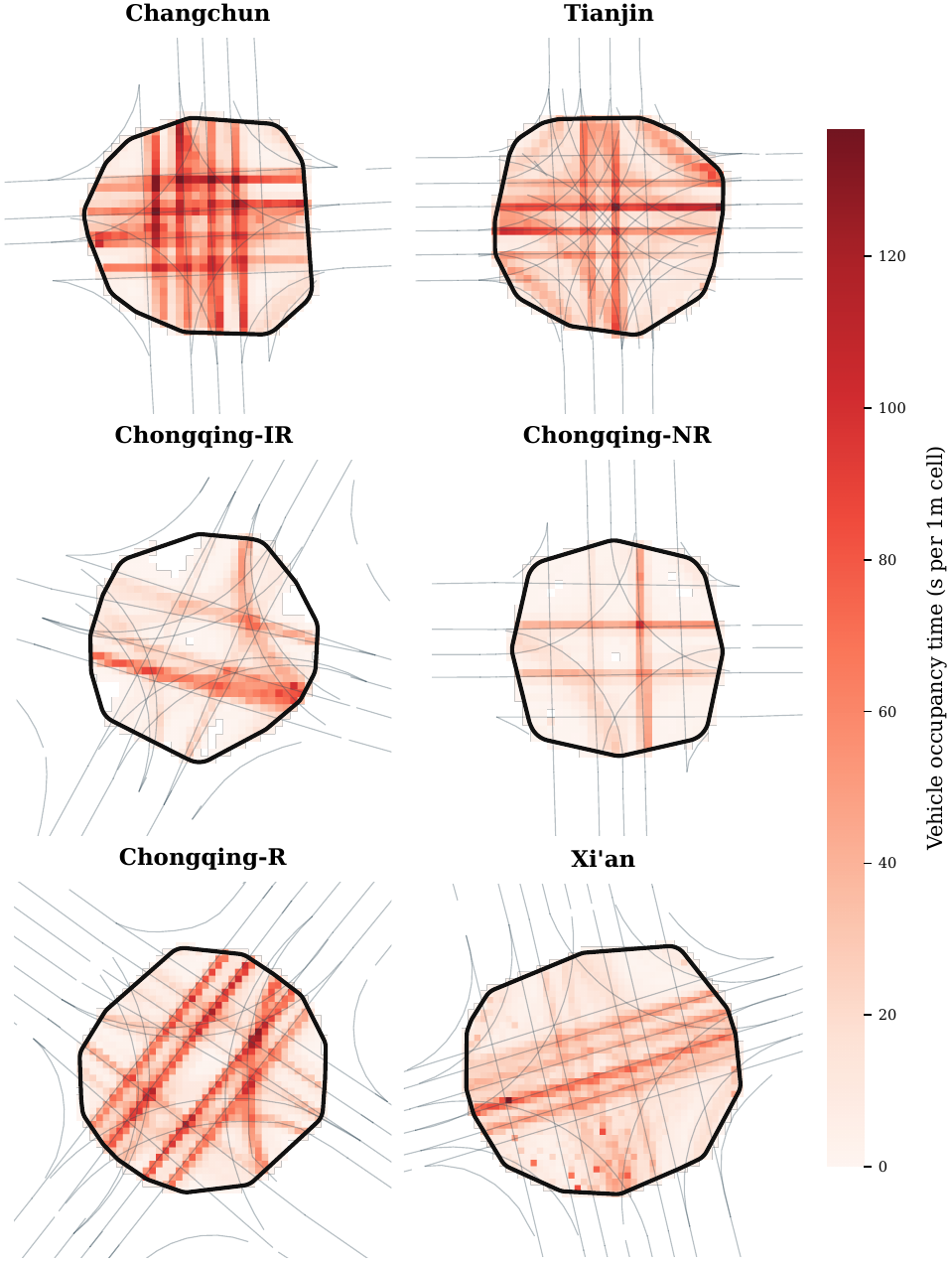}
    \caption{Spatiotemporal vehicle occupancy density for all six intersections. Each panel is cropped to the active core ROI to emphasize the recurrent accumulation region.}
    \label{fig:occupancy_density}
\end{figure}

Figure~\ref{fig:occupancy_density} shows that the high-exposure region takes different shapes across domains: Changchun is more concentrated, Xi'an is broad and diffuse, and Chongqing-R accumulates along curved or weaving movement corridors. The Xi'an panel also complements the VRU analysis in Figure~\ref{fig:xasl_trapped_vru}: vehicles occupy a wide central region for long durations, while VRUs traverse long unprotected paths through the same intersection core. This coupling creates a natural stress test for prediction and planning because the ADS must reason about recurrent vehicle congestion and low-speed dynamic obstacles in the same spatial region.

\subsection{Interaction Topology and Conflict Patterns}

\subsubsection{Multi-Agent Interaction Degree}

Multi-agent interaction degree characterizes the topological structure of multi-agent interactions at intersections, which is a primary source of complex risk relations and a core component of ADS safety test scenarios. Interaction components are constructed by connecting agents within 15 m proximity whose predicted trajectories may conflict over a 5-second horizon. Each connected component forms an interaction group, with its graph degree quantifying the number of agents engaged in the same local interactive game. This definition prioritizes interaction complexity over immediate collision risk, capturing high-challenge scenarios for prediction and planning modules.

\begin{figure}[t]
    \centering
    \includegraphics[width=0.88\columnwidth]{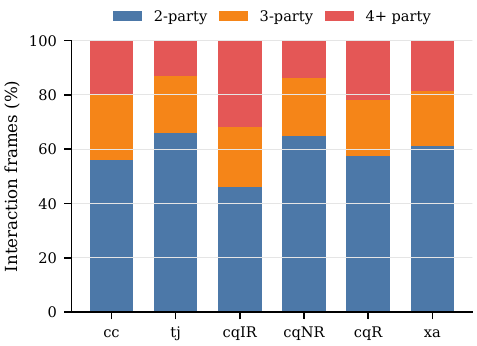}
    \caption{Distribution of interaction degree by intersection. }
    \label{fig:interaction_degree}
\end{figure}

Figure~\ref{fig:interaction_degree} presents the interaction degree distribution at individual intersections. Chongqing-IR exhibits the highest interaction complexity: 54.04\% of component frames involve at least three agents, and 31.99\% involve four or more, indicating the richest set of potential multi-vehicle conflict patterns across all sites. In contrast, Tianjin is dominated by pairwise interactions, though 34.18\% of frames still feature $N\geq3$ agents. Its predominantly two-party interaction profile makes it well suited for safety analysis and testing of canonical scenarios such as left-turn through conflicts. Overall, the diverse multi-agent interaction behaviors captured in SinD 2.0 support test scenario development across varying levels of interaction complexity.

\subsubsection{Interaction duration}
Interaction duration measures how long an interaction component remains active before the involved agents separate or resolve their local game. Unlike instantaneous proximity or surrogate collision risk, this metric captures temporal persistence: a long interaction requires an ADS to maintain prediction consistency, update right-of-way beliefs, and avoid premature or oscillatory planning decisions over multiple seconds.

\begin{table}[t]
\centering
\caption{Interaction-Duration Statistics Across Intersections}
\label{tab:interaction_duration_summary}
\resizebox{\columnwidth}{!}{%
\begin{tabular}{lrrrr}
\toprule
Intersection & Median (s) & P90 (s) & Long game $\geq$5 s & $N\geq3$ ratio \\ \midrule
Changchun (cc) & 2.00 & 7.58 & 19.42\% & 44.01\% \\
Tianjin (tj) & 2.70 & 10.10 & 27.39\% & 34.18\% \\
Chongqing-IR (cqIR) & 3.60 & 21.27 & 39.34\% & 54.04\% \\
Chongqing-NR (cqNR) & 3.20 & 15.70 & 31.39\% & 35.23\% \\
Chongqing-R (cqR) & 3.80 & 8.90 & 32.63\% & 42.35\% \\
Xi'an (xa) & 2.50 & 9.80 & 28.31\% & 38.70\% \\
\bottomrule
\end{tabular}%
}
\end{table}

Table~\ref{tab:interaction_duration_summary} shows clear duration heterogeneity across the six domains. Chongqing-IR has the most persistent interactions, with a P90 duration of 21.27 s, a 39.34\% long-game ratio, and the highest $N\geq3$ ratio. Chongqing-NR also exhibits extended interactions, with a P90 duration of 15.70 s, despite having a lower multi-agent ratio than Chongqing-IR. In contrast, Changchun has a shorter P90 duration of 7.58 s, indicating that its dense interactions are more frequently resolved as shorter local encounters.

These differences are consistent with the underlying domain structure. Skewed or compressed intersections can keep agents in overlapping negotiation regions for longer periods, while grid-like sites more often produce short crossing or yielding episodes. The testing implication is that long games stress a different capability from near-miss avoidance: the ADS must preserve progress while repeatedly updating predictions under ambiguous right-of-way, instead of executing a single evasive maneuver.

\begin{figure}[t]
    \centering
    \includegraphics[width=0.9\columnwidth]{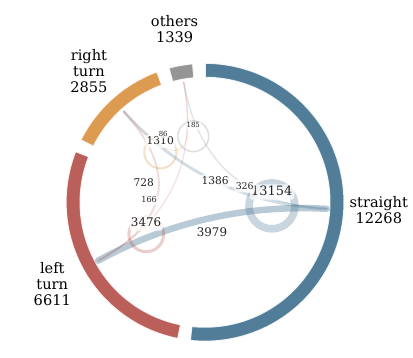}
    \caption{Conflict-mode composition of SinD 2.0.}
    \label{fig:conflict_patterns}
\end{figure}

\subsubsection{Conflict-mode signatures}

Figure~\ref{fig:conflict_patterns} depicts the interaction conflict patterns in SinD 2.0, categorized by driving intent. Overall, through-through, through-left-turn, and left-turn-left-turn conflicts are the predominant types, which typically introduce elevated collision risks in intersection scenarios. Conflict pattern distributions vary across intersections, and these patterns in turn define distinct interaction grammars and corresponding safety challenges. Table~\ref{tab:representative_conflict_modes} summarizes the representative conflict mode signature for each intersection, demonstrating that cross-site discrepancies manifest not only quantitatively but also in the qualitative structure of interactions. Specifically, Changchun and Tianjin both retain a regular grid structure, but their signatures are not identical: Changchun combines crossing with a high weaving component, whereas Tianjin has the strongest merging component among the selected sites. Chongqing-IR is a qualitatively different domain because crossing conflicts almost disappear, while turning and weaving dominate the interaction set. Chongqing-NR is more balanced, with crossing and weaving shares both near 37\%, showing that not all Chongqing sites collapse into the same conflict grammar. Chongqing-R is more specialized, with weaving or parallel competition becoming the primary conflict mode, making it a stress test for negotiation in curved or roundabout-like flows. Xi'an sits between the grid and mountain-city regimes: it preserves a non-trivial crossing share but also has strong weaving and mixed-other components, matching the VRU spillover evidence in Figure~\ref{fig:xasl_trapped_vru}.
This heterogeneity renders SinD 2.0 well suited for establishing cross-domain test benchmarks and safety-critical scenario libraries.

\begin{table*}[t]
\centering
\caption{Representative Conflict-Mode Signatures Across Intersections}
\label{tab:representative_conflict_modes}
\begin{tabular}{lrrrrl}
\toprule
Intersection & Crossing & Merging & Weaving & Turning & Dominant Cross-Domain Signature \\ \midrule
Changchun (cc) & 44.39\% & 4.91\% & 32.81\% & 15.42\% & grid-like crossing plus large-vehicle mixing \\
Tianjin (tj) & 46.43\% & 15.86\% & 19.57\% & 16.52\% & large grid intersection with crossing/merging exposure \\
Chongqing-IR (cqIR) & 3.36\% & 15.31\% & 38.10\% & 35.97\% & skewed topology with long turning and merging games \\
Chongqing-NR (cqNR) & 36.99\% & 6.31\% & 36.74\% & 18.69\% & narrow regularized flow with crossing-weaving balance \\
Chongqing-R (cqR) & 1.96\% & 1.37\% & 66.04\% & 24.26\% & compressed roundabout-like weaving domain \\
Xi'an (xa) & 29.91\% & 6.84\% & 33.94\% & 16.86\% & mixed VRU spillover and vehicle competition \\
\bottomrule
\end{tabular}
\end{table*}

Overall, this analysis shows that SinD 2.0 is not merely a larger collection of intersection trajectories, but a cross-domain dataset with measurable diversity in geometry, motion, infrastructure, and interaction grammar. Large un-channelized areas prolong exposure, skewed and compressed layouts create persistent multi-agent games, heterogeneous motion envelopes reveal different dynamic boundaries, and conflict-mode shifts expose qualitatively different planning problems. These differences more faithfully reflect the complexity of real-world urban traffic, where an ADS must handle not only isolated critical events but also long negotiations, recurrent congestion, mixed road users, and locally specific right-of-way ambiguities.

From a safety-testing perspective, each domain contributes a distinct pressure point. Long games challenge forecasting models to remain stable over extended interaction horizons; weaving and parallel-competition patterns stress planning under curved or compressed feasible spaces; and infrastructure-induced VRU exposure stresses perception, prediction, and decision making when vulnerable road users remain inside vehicle conflict zones. By combining these complementary stressors, SinD 2.0 provides a more comprehensive and challenging platform for ADS safety research, supporting evaluation protocols that move beyond single-site performance toward broader real-world generalization.

\section{Safety-Critical Scenario Mining and Semantic Annotation}\label{sec:semantic}

SOTIF testing relies on mining scenarios with specific trigger conditions from naturalistic traffic data, including traffic rule violations, close-proximity conflicts, agents emerging from visual occlusions, and narrow drivable areas. To this end, SinD 2.0 constructs a semantic scene layer upon continuous trajectory data, high-definition (HD) maps, and traffic signal states, converting raw naturalistic data into structured, searchable test cases. Each case is characterized by a well-defined time window, an ego vehicle, interacting agents, and event-specific attributes. This layer enables test engineers to retrieve scenarios based on targeted trigger conditions, thereby substantially improving the efficiency of SOTIF testing.

\subsection{Hierarchical Semantic Labeling Framework}

We organize the semantic labels as a hierarchical scenario schema. Each scenario contains common fields shared by all labels, including a unique scenario ID, the source recording, a frame-level time window, ego-agent ID, semantic tags, and source metadata. Event-specific fields are then attached under the \textit{semantics} field. This design allows different risk types to share the same retrieval and testing interface while preserving the additional information needed by each scenario family.

\begin{figure}[t]
    \centering
    \includegraphics[width=\columnwidth]{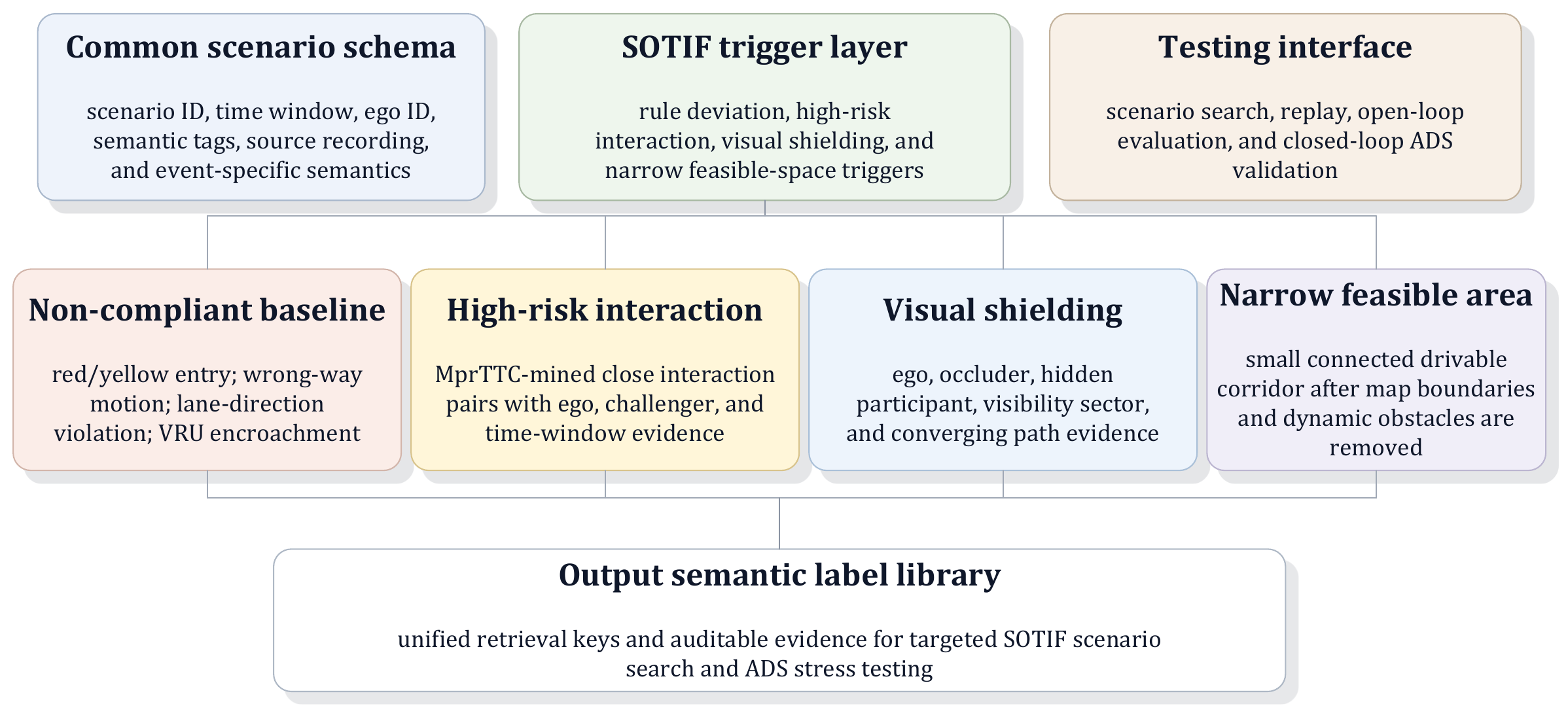}
    \caption{Hierarchical semantic scenario labeling framework in SinD 2.0. A shared schema stores common scenario metadata, while event-specific semantic fields describe non-compliance, high-risk interaction, visual shielding, and narrow feasible-area triggers.}
    \label{fig:semantic_taxonomy}
\end{figure}

Figure~\ref{fig:semantic_taxonomy} summarizes the taxonomy. The current full semantic label library contains 53,901 scenario-level records: 31,005 high-risk MprTTC scenarios, 1,286 visual-shielding scenarios, and 21,610 narrow-feasible-area scenarios. In addition, the structured violation and non-compliance miner provides a rule-based baseline layer with 25,966 event records, including 19,550 vehicle-side violation events and 6,416 VRU-side non-compliance events. These labels are not intended to replace continuous trajectory learning or to serve as manually certified ground truth. Instead, they provide an auditable bridge from naturalistic data to targeted ADS tests: a user can query a semantic label, inspect its stored evidence, instantiate its time window and agents, and replay or simulate the corresponding interaction in the testing toolchain.

\subsection{Violations and Non-Compliant Behavior}

Traffic-rule deviations form the background risk distribution of real-world intersections. Even before considering complex triggers such as occlusion, an ADS operating in mixed urban traffic cannot assume that surrounding agents are rule-based. This issue is especially important in Chinese urban intersections, where non-motorized vehicles and pedestrians frequently deviate from nominal signal, lane, or crosswalk rules, and their behavior can interact with motor-vehicle-side violations.

We therefore mine structured non-compliant behaviors from HD maps, signal states, crosswalk geometry, and trajectories. The motor-vehicle-side labels include red-light running, yellow-light entry, potential wrong-way behavior, and pre-entry solid-line crossing near the stop line. The non-motorized and pedestrian-side labels include non-motorized red-light running, potential two-wheeler wrong-way behavior, and pedestrian outside-crosswalk path exposure.

\begin{figure}[t]
    \centering
    \includegraphics[width=0.9\columnwidth]{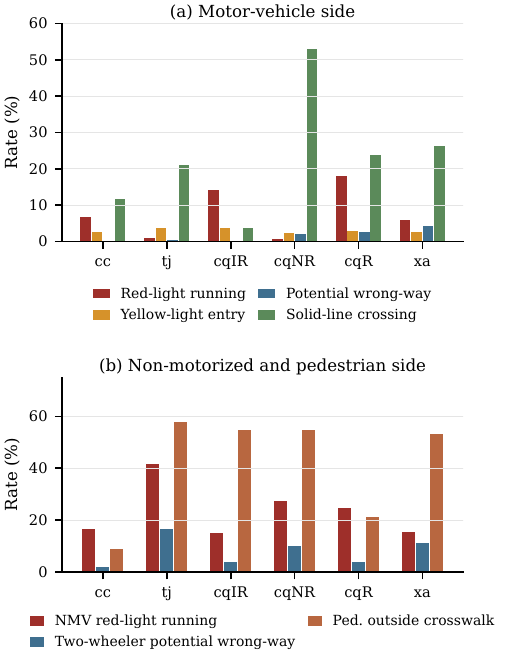}
    \caption{Structured violation and non-compliance rates across intersections. (a) Motor-vehicle-side red-light running, yellow-light entry, potential wrong-way behavior, and pre-entry solid-line crossing rates. Red-light running is measured at the corresponding stop line, and solid-line crossing is measured before stop-line entry. (b) Non-motorized red-light running, potential two-wheeler wrong-way behavior rate, and pedestrian aggregate outside-crosswalk path ratio. Pedestrian ratios are shown only where crosswalk polygons provide a valid denominator.}
    \label{fig:violation_rates}
\end{figure}

Figure~\ref{fig:violation_rates} shows that structured rule deviations are strongly domain-dependent. On the motor-vehicle side, Chongqing-R has the highest red-light running rate (18.06\%), while Chongqing-IR is the second highest (14.04\%). The pre-entry solid-line crossing rate is particularly high at Chongqing-NR (52.86\%), followed by Xi'an (26.36\%), Chongqing-R (23.66\%), and Tianjin (20.98\%). Xi'an has the highest motor-vehicle potential wrong-way behavior rate (4.28\%). On the non-motorized and pedestrian side, Tianjin has the highest non-motorized red-light running rate (41.71\%) and the highest potential two-wheeler wrong-way behavior rate (16.47\%). For pedestrian path behavior, the aggregate outside-crosswalk path ratio is highest in Tianjin (57.70\%), followed by Chongqing-IR (54.86\%), Chongqing-NR (54.78\%), Xi'an (53.34\%), Chongqing-R (21.32\%), and Changchun (8.94\%) among locations with valid crosswalk denominators. These statistics provide the non-compliant baseline for safety validation.

\begin{figure*}[t]
    \centering
    \includegraphics[width=0.96\textwidth]{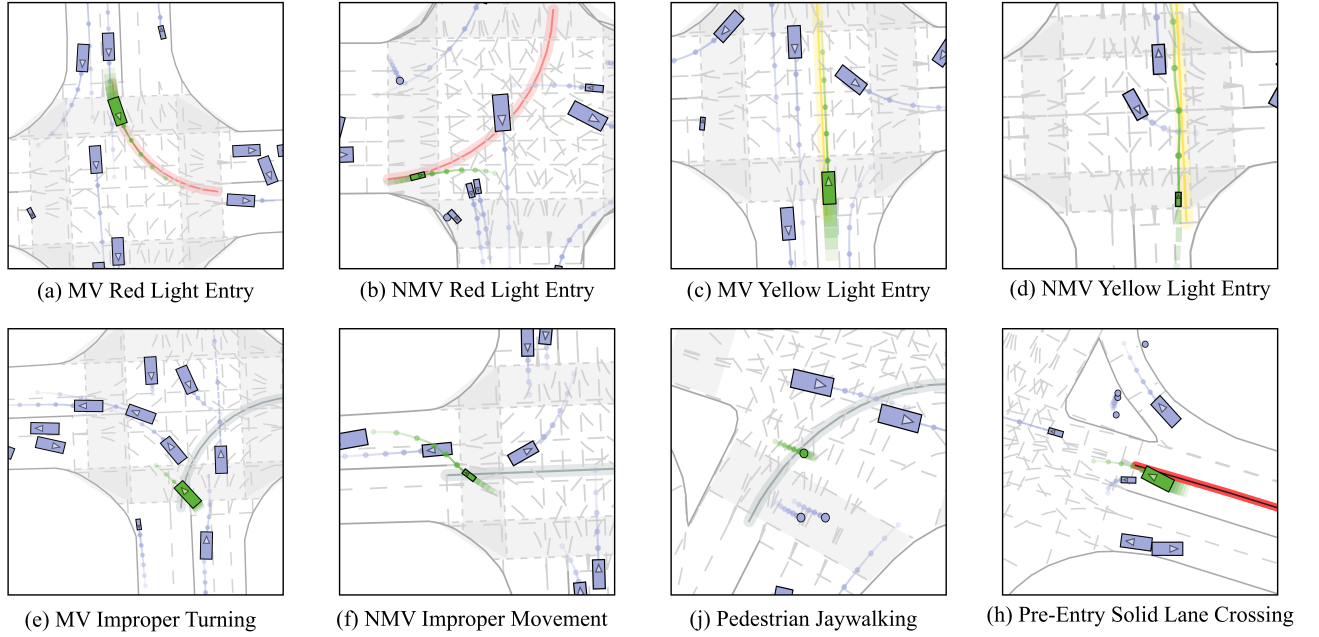}
    \caption{Representative structured violation and non-compliance examples. The montage shows motor-vehicle (MV) and non-motorized-vehicle (NMV) red-light entry, MV and NMV yellow-light entry, MV improper turning, NMV improper movement, pedestrian outside-crosswalk walking, and pre-entry solid-lane crossing.}
    \label{fig:violation_cases}
\end{figure*}

Figure~\ref{fig:violation_cases} provides representative examples. The extraction procedure is rule-based. For signal-related labels, trajectories are bound to lane-level traffic-light states, and an event is emitted when the participant crosses the corresponding stop-line or enters the core ROI under a red or yellow phase. For maneuver-compliance labels, the stable approach lane before ROI entry determines the allowed movement set, while the observed entry-exit geometry determines the realized movement. For wrong-way labels, the heading of a moving participant is compared with the nearest Lanelet2 centerline outside the core ROI. For pedestrian outside-crosswalk behavior, path length is accumulated inside and outside parsed crosswalk polygons, and a trajectory is marked when most of its path is outside the legal crossing area. Static or long-stationary objects are filtered before labeling to avoid confusing parked or roadside objects with dynamic non-compliance.

\subsection{High-Risk Interaction Events}

The second semantic family targets high-risk interactions. Unlike rule-deviation labels, these scenarios are selected by interaction criticality rather than by legal compliance. Following the surrogate-safety-measure tradition of TTC-based conflict screening~\cite{mahmud2017application}, we use MprTTC to measure the earliest predicted collision time between two agents under short-horizon motion rollout.

\begin{figure}[t]
    \centering
    \includegraphics[width=0.94\columnwidth]{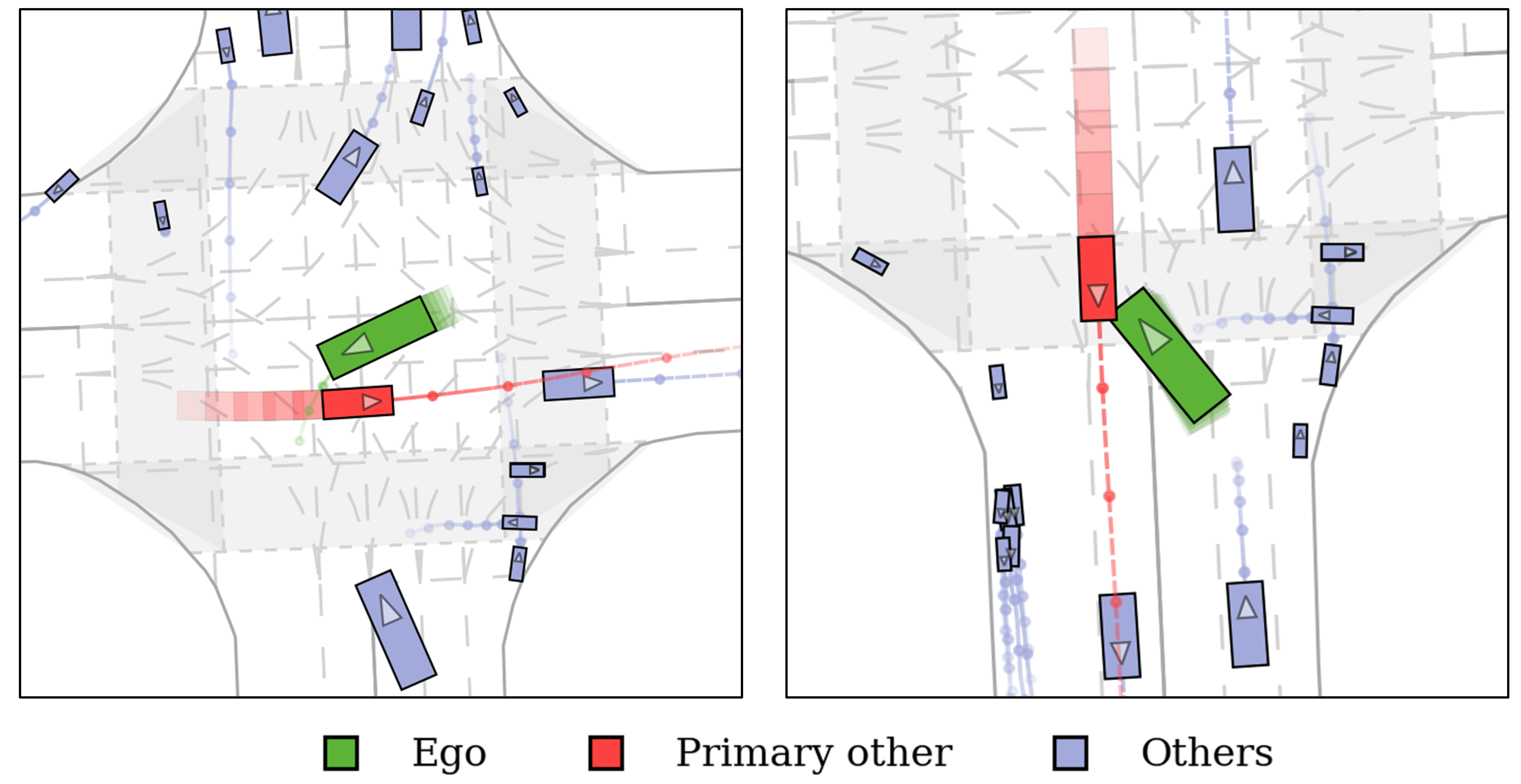}
    \caption{Representative high-risk interactions mined by MprTTC. The green ego vehicle and red primary other are shown with recent motion history and short-horizon reachable occupancy, while surrounding traffic provides the local interaction context.}
    \label{fig:mprttc_case}
\end{figure}

For an agent pair $(i,j)$ at frame $t$, let $\mathcal{B}_i(t+\Delta)$ and $\mathcal{B}_j(t+\Delta)$ denote the predicted occupied shapes after $\Delta$ frames, obtained by rolling forward the current position, velocity, acceleration, and heading. The pairwise MprTTC score is defined as
\begin{equation}
\begin{aligned}
\mathrm{MprTTC}_{ij}(t)
&=\min_{\Delta\in\{1,\ldots,H\}} \Delta\,\delta t,\\
\mathrm{s.t.}\quad
&\mathcal{B}_i(t+\Delta)\cap\mathcal{B}_j(t+\Delta)\neq\emptyset ,
\end{aligned}
\end{equation}
where $\delta t$ is the frame period. The score is set to $+\infty$ when no predicted overlap occurs within horizon $H$. A high-risk interaction event is then identified by
\begin{equation}
\begin{aligned}
\mathbb{I}_{\mathrm{HR}}(i,j,t)=
\mathbf{1}\left[
\right.&\mathrm{MprTTC}_{ij}(t)\leq \tau_{\mathrm{mpr}}\\
&{}\land d_{ij}(t)\leq d_{\max}\\
&{}\left.\land (i\in\mathcal{V}_{m}\lor j\in\mathcal{V}_{m})
\right],
\end{aligned}
\end{equation}
where $d_{ij}(t)$ is the current pair distance, $d_{\max}$ is the candidate-pair pruning radius, and $\mathcal{V}_{m}$ denotes motor vehicles. Consecutive positive frames are merged into one scenario and stored with the ego ID, challenger ID, time window, event frame, and minimum MprTTC value. Figure~\ref{fig:mprttc_case} illustrates two mined cases where the ego and the primary other approach a near-overlapping future occupancy region from different directions.

\subsection{Visual-Shielding Scenarios}

Visual shielding describes scenarios where risk stems from delayed perception, in which an agent is occluded by a vehicle or obstacle before abruptly entering the ego vehicle's path. These scenarios are critical for SOTIF: an ADS may respond correctly to visible objects yet fail to anticipate occluded, converging agents. Such abrupt exposure imposes severe challenges to ADS perception and planning modules, as agents can enter the ego's field of view and trigger conflicts with insufficient reaction margin.

For each frame, the ego vehicle's visibility sector is defined by a radius $R$ and field-of-view angle $\theta$. A target participant $i$ is considered within the nominal visible region if
\begin{equation}
\alpha_i = \arccos\left(\frac{\vec{v}_{e} \cdot (\vec{p}_{i}-\vec{p}_{e})}{\|\vec{v}_{e}\| \|\vec{p}_{i}-\vec{p}_{e}\|}\right) \leq \frac{\theta}{2},
\quad
d_i=\|\vec{p}_{i}-\vec{p}_{e}\| \leq R,
\end{equation}
where $\vec{p}_{e}$ and $\vec{v}_{e}$ denote the ego position and velocity. Occlusion is then determined by checking whether the angular span of a closer occluding participant covers the boundary rays of the target participant from the ego viewpoint. Finally, the hidden participant must have an approaching trend and a future path intersection with the ego trajectory. The approaching trend is evaluated using cross products between the ego direction, target direction, and relative position:
\begin{equation}
\text{Cross}_{\text{pos}}=\vec{d}_{e}\times(\vec{p}_{i}-\vec{p}_{e}), \qquad
\text{Cross}_{\text{dir}}=\vec{d}_{e}\times\vec{d}_{i}.
\end{equation}
Opposite signs of these two quantities indicate converging motion from the left or right side.

\begin{algorithm}[t]
\vspace{2mm}
\caption{Rule-Based Visual-Shielding Scenario Extraction Procedure}
\label{alg:visual_shielding}
\begin{algorithmic}[1]
\FOR{each frame $t$ in each recording}
    \STATE Compute the ego visibility sector using $(R,\theta)$.
    \FOR{each target participant $i$ within the sector}
        \STATE Search closer participants $j$ whose angular span covers $i$.
        \IF{$i$ is occluded by $j$}
            \STATE Test whether $i$ has a converging motion trend.
            \STATE Predict short-horizon paths of ego and $i$.
            \IF{paths intersect within time tolerance $\epsilon$}
                \STATE Save scenario $(e,j,i,t)$ with occlusion semantics.
            \ENDIF
        \ENDIF
    \ENDFOR
\ENDFOR
\end{algorithmic}
\vspace{2mm}
\end{algorithm}

\begin{figure}[t]
    \centering
    \includegraphics[width=0.94\columnwidth]{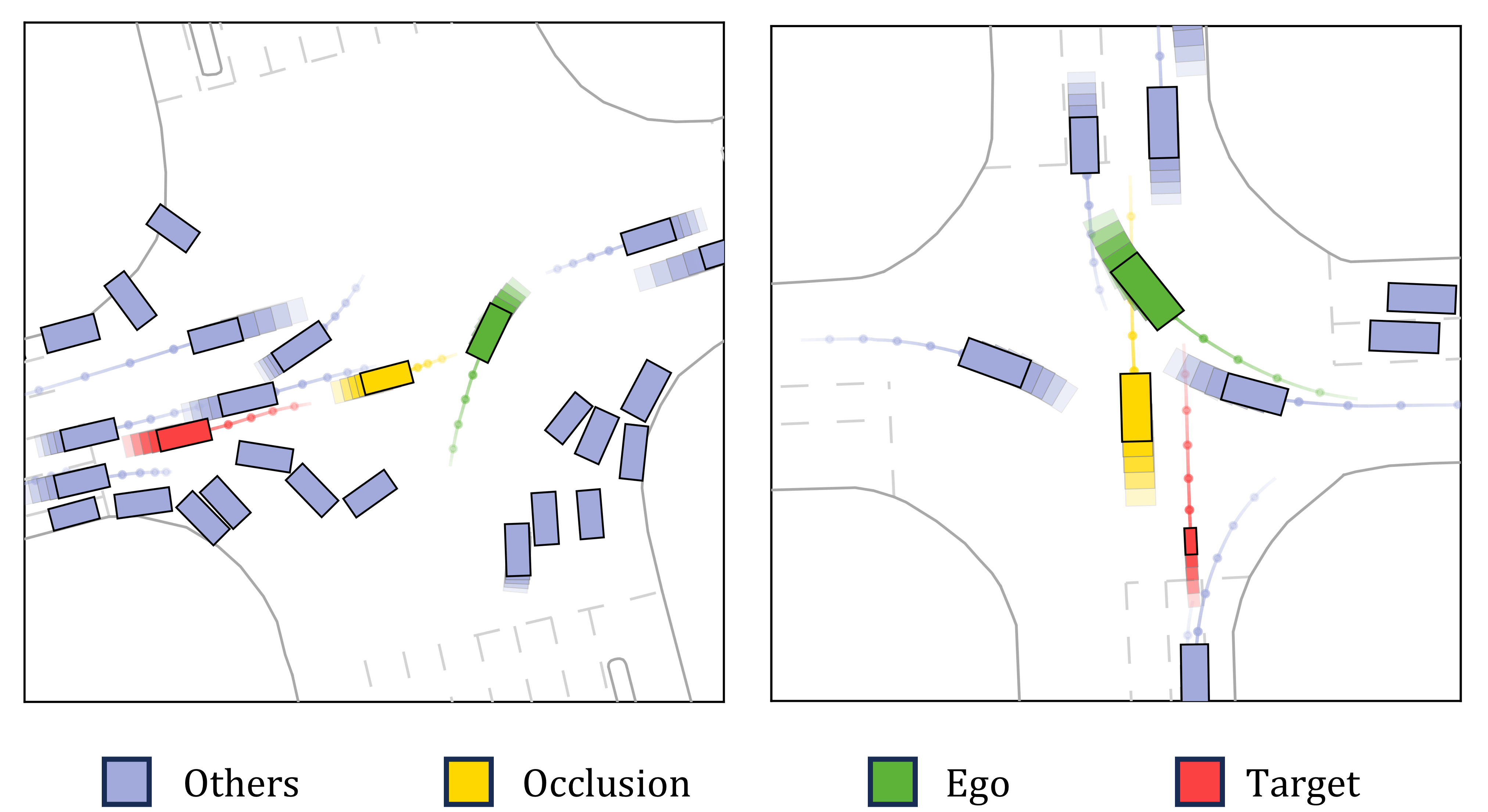}
    \caption{Representative visual-shielding scenarios. The yellow occluding participant blocks the line of sight between the green ego vehicle and the red target, while the target's trajectory later converges with the ego's path.}
    \label{fig:occlusion_placeholders}
\end{figure}

Algorithm~\ref{alg:visual_shielding} outlines the extraction process. The current semantic label library contains 1,286 visual-shielding scenarios. Figure~\ref{fig:occlusion_placeholders} shows representative cases in which an occluding traffic participant shields a later-conflicting target, turning an otherwise ordinary crossing or turning interaction into a perception-limited SOTIF trigger.

\subsection{Narrow Feasible-Area Scenarios}

The fourth semantic family focuses on narrow feasible-area events. In car-following or highway scenarios, urgency is often characterized by the distance to a specific leading vehicle. At intersections, the more relevant constraint can be the remaining feasible drivable corridor itself. An ego vehicle may face multiple surrounding agents, curbs, and lane boundaries without a single obvious lead vehicle. When the feasible area becomes narrow, the risk can increase directly with the ego's initial speed because the planner has less lateral and longitudinal space to recover.

We define the feasible region in an ego-centric grid placed in front of the vehicle. At each frame, the grid is first clipped by map boundaries and occupied cells induced by inflated surrounding traffic participants. A depth-first search is then performed from the ego-adjacent cells along the forward direction, with lateral expansion constrained by the vehicle's kinematically feasible motion range. This produces a connected drivable component that is reachable from the current ego state. The forward depth of this component is used as the bottleneck measurement, and a narrow feasible-area event is emitted when the maximum reachable depth remains below a threshold for consecutive frames.

\begin{algorithm}[t]
\vspace{2mm}
\caption{Narrow Feasible-Area Scenario Extraction Procedure}
\label{alg:narrow_feasible}
\begin{algorithmic}[1]
\FOR{each ego trajectory in the dataset}
    \FOR{each frame $t$ in the trajectory}
        \STATE Build a forward ego-centric drivable grid from the HD map.
        \STATE Mark cells blocked by map boundaries and inflated participants.
        \STATE Run forward depth-first search from ego-adjacent cells.
        \STATE Retain cells connected to the ego under kinematic reachability.
        \STATE Compute maximum forward reachable depth $D_{\text{free}}(t)$.
        \IF{$D_{\text{free}}(t) < D_{\min}$ for consecutive frames}
            \STATE Mark frame $t$ as a narrow-feasible-area frame.
        \ENDIF
    \ENDFOR
    \STATE Merge consecutive marked frames into scenario windows.
\ENDFOR
\end{algorithmic}
\vspace{2mm}
\end{algorithm}

\begin{figure}[t]
    \centering
    \includegraphics[width=0.94\columnwidth]{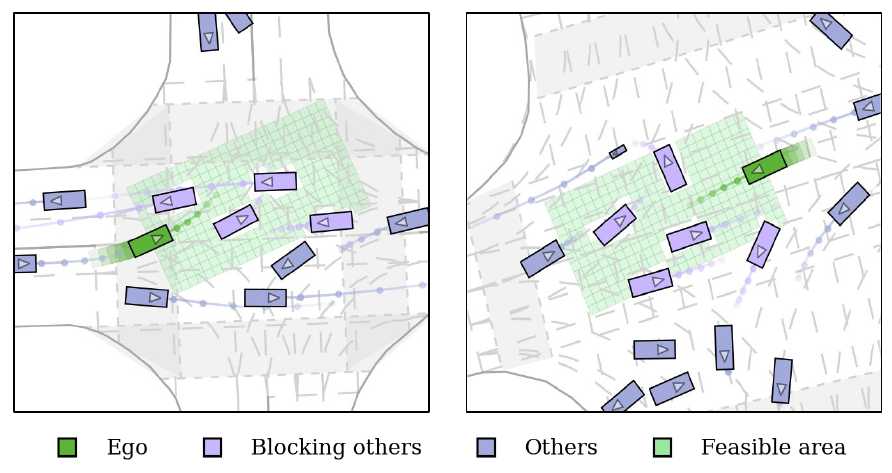}
    \caption{Representative narrow feasible-area scenarios. Green cells denote the reachable grid component connected to the ego vehicle after map boundaries and inflated dynamic obstacles are removed; gray cells denote blocked or unreachable front-grid cells, and purple participants are blocking others that compress the available maneuver space.}
    \label{fig:narrow_feasible_case}
\end{figure}

Algorithm~\ref{alg:narrow_feasible} summarizes the extraction procedure, and Figure~\ref{fig:narrow_feasible_case} shows representative examples where the ego remains in a small connected drivable component surrounded by blocking participants. The implementation records the per-frame feasible-area values, the minimum area within the scenario window, the ego identifier, and the blocking participants used to subtract dynamic obstacles. The full semantic label library contains 21,610 narrow-feasible-area scenarios. These labels are particularly useful for planner evaluation because the same geometric bottleneck can be replayed at different ego initial speeds or with different policy settings, turning naturalistic data into a structured feasible-space boundary test.

\paragraph{Auditability and limitations.}
All semantic labels are generated by deterministic rules from trajectories, maps, SPaT states, and stored geometric evidence. This design makes each label reproducible and inspectable, but it also means that the labels inherit map, signal-binding, and trajectory-estimation errors. We therefore use the semantic layer as a queryable scenario-mining index whose thresholds and evidence can be inspected and revised, rather than as a finalized benchmark of human-verified annotation accuracy. In the current draft, we report label definitions, stored evidence fields, representative cases, and aggregate statistics; a larger manual precision/recall audit and threshold-sensitivity study are left for the dataset release validation.

\section{Experiments and Toolchain Validation}\label{sec:experiments}
This section presents the safety-oriented toolchain of SinD 2.0 and validates its effectiveness for autonomous driving safety testing via evaluations on various ADS algorithms and systems. The experiments cover trajectory prediction, open-loop testing and closed-loop testing. We further verify the efficacy of the semantic labels proposed in Section \ref{sec:semantic} within open-loop tests. The results demonstrate that SinD 2.0 functions not only as a trajectory dataset but also as a comprehensive platform supporting diverse testing paradigms, enabling researchers to efficiently identify and analyze potential safety hazards of autonomous driving systems.

\subsection{Testing Toolchain and Evaluation Modes}

The released toolchain is built on \emph{trajdata}, a unified interface for
multiple human trajectory datasets \cite{ivanovic2023trajdata}. Its shared data
abstraction and dataset adapters enable the same evaluation workflow to be
reused across SinD 2.0 and other supported trajectory datasets. For SinD 2.0,
the toolchain converts a trajectory segment or a semantic label into an
executable test case. Each test case specifies the intersection, scene index,
ego agent, initial time, evaluation horizon, ego policy, non-ego policy, and
metrics to be logged. Semantic labels can be used directly as test selectors,
so users can instantiate MprTTC, visual-shielding, or narrow-feasible-area
cases without manually searching the raw recordings.

For experiments, three evaluation modes are used.
\begin{itemize}
    \item \textbf{Offline trajectory prediction evaluation:} A track is split once into history and future, the model predicts future motion, and ADE/FDE/MR are computed against the held-out future.
    \item \textbf{Open-loop testing:} Replacing the ego trajectory with a policy output while all non-ego participants replay the recorded trajectories.
    \item \textbf{Closed-loop testing:} The ego policy still controls the target agent, but nearby non-ego agents are interactively updated by a
data-driven reactive model. In our implementation, non-ego agents within a
25\,m neighborhood are eligible for Diffuser-based reactive control, capped at the five
closest forward relevant agents, with a 5-frame non-ego update interval. This
setting is closer to safety validation because background traffic can react to
the ego behavior rather than remaining a static replay.
\end{itemize}

\subsection{Experiment Setup and Metrics}

Unless otherwise noted, the experiments in this section use fixed released
checkpoints. The evaluated policies and predictors are:
\begin{itemize}
    \item \textbf{RiskIDM~\cite{liHighdimensionalFunctionalBoundaries2025}}: a rule-based intersection controller extending IDM with a risk-aware interaction module.
    \item \textbf{ASAPRL~\cite{asaprl}}: a hierarchical learning-based driving policy combining imitation learning and reinforcement learning.
    \item \textbf{QCNet~\cite{qcnet}}: a vector-map-based trajectory prediction model evaluated with a released public checkpoint.
    \item \textbf{Diffuser~\cite{diffuser}}: a diffusion-based trajectory generation model trained on SinD 2.0 with Bicycle-dynamics compatibility enabled during evaluation.
\end{itemize}
For prediction-only evaluation, QCNet uses 50 history frames and 60 future
frames, while Diffuser uses 31 history frames, a 52-frame prediction horizon, and ten evaluation samples.
ASAPRL and RiskIDM use the current simulation-toolchain implementations and
checkpoints documented in the released scripts. The present results should
therefore be read as toolchain-integrated transfer tests rather than as a fully
standardized leaderboard benchmark.

For offline prediction evaluation, the accuracy metrics are reported, including ADE, FDE, and MR. For open-loop and closed-loop testing, the metrics include interaction-risk indicators (MinTTC, AveTTC, MRD, ARD), and behavior validity indicators (collision, off-road, wrong-way, red-light, and lane-direction rule violations). MinTTC is computed using the MprTTC-style forward bicycle-model rollout, where smaller values indicate more critical interactions. The anomaly rate reports the fraction of completed test runs whose trajectory deviation becomes excessively large, defined in the evaluation scripts as ADE $>10\,\mathrm{m}$ or FDE $>50\,\mathrm{m}$. It is therefore used to flag metric-level rollout anomalies rather than ordinary collision, off-road, or traffic-rule violation events.

\begin{table*}[t]
\centering
\caption{Benchmark protocol of experiments.}
\label{tab:benchmark_protocol}
\footnotesize
\setlength{\tabcolsep}{3pt}
\begin{tabular}{@{}p{0.14\textwidth}p{0.17\textwidth}p{0.19\textwidth}p{0.19\textwidth}p{0.23\textwidth}@{}}
\toprule
Mode                        & Policy / model         & Input context                 & Horizon / sampling                              & Metrics                               \\ \midrule
\multirow{2}{*}{Prediction} & QCNet                  & Track history, vector map     & 50 history frames, 60 future frames, 10 samples & One-shot ADE/FDE/MR                   \\
                            & Diffuser               & Track history, local context  & 31 history frames, 52-frame horizon, 10 samples & One-shot ADE/FDE/MR                   \\
Dataset-open-loop           & RiskIDM, ASAPRL, QCNet & Recorded non-ego replay       & Scenario-specific horizon                       & Collision, off-road, and rule metrics \\
Reactive closed-loop        & RiskIDM, ASAPRL, QCNet & Reactive non-ego neighborhood & 25 m neighborhood relevant agents               & Collision, off-road, and rule metrics \\ \bottomrule
\end{tabular}
\end{table*}

\subsection{Offline Trajectory Prediction Benchmark}

Table~\ref{tab:benchmark_protocol} summarizes the evaluation protocol, and
Table~\ref{tab:prediction_benchmark} reports the prediction
results on SinD 2.0, Argoverse 2, and nuScenes.

The results reveal that QCNet achieves superior performance on SinD 2.0 and Argoverse 2, indicating stronger generalization under varying map layouts across these datasets. On the nuScenes validation split, Diffuser yields lower top-1 displacement error than QCNet, yet it suffers degraded oracle metrics and higher miss rates. This implies that the current Diffuser configuration occasionally generates plausible single trajectories, while its overall candidate quality and recovery capability for hard scenarios remain constrained.

\begin{table*}[t]
\centering
\caption{Offline prediction-only benchmark across SinD 2.0, Argoverse 2 validation, and nuScenes validation. ADE/FDE are in meters; MR is the miss rate with the 2\,m FDE threshold.}
\label{tab:prediction_benchmark}
\footnotesize
\begin{tabular}{llrrrrrr}
\toprule
Dataset                       & Model    & top1 ADE & top1 FDE & minADE & minFDE & MR   \\ \midrule
\multirow{2}{*}{SinD 2.0}     & QCNet    & 2.60     & 6.57     & 0.78   & 1.52   & 11.2 \\
                              & Diffuser & 3.88     & 10.13    & 3.88   & 10.13  & 87.6 \\ \midrule
\multirow{2}{*}{AV2 val}      & QCNet    & 2.97     & 8.04     & 1.21   & 2.77   & 34.3 \\
                              & Diffuser & 4.45     & 11.67    & 4.45   & 11.66  & 93.5 \\ \midrule
\multirow{2}{*}{nuScenes val} & QCNet    & 4.25     & 11.00    & 1.54   & 3.67   & 37.5 \\
                              & Diffuser & 3.13     & 8.52     & 3.13   & 8.51   & 77.3 \\ \bottomrule
\end{tabular}
\end{table*}

\subsection{Dataset Open-Loop Evaluation on Semantic Scenarios}

To validate the effectiveness of the extracted safety-critical scenarios, we conduct an open-loop evaluation on raw naturalistic dataset and extracted semantic scenarios. The raw naturalistic dataset serves as the nominal baseline. As demonstrated in Table \ref{tab:semantic_open_loop}, the rule-based RiskIDM exhibits a conservative driving profile in raw naturalistic scenarios with a minimal collision rate of 3.0\%. Conversely, data-driven policies including ASAPRL and QCNet display higher baseline collision and anomaly rates. This performance discrepancy is a well-documented phenomenon in dataset-open-loop testing, where the lack of reactive background traffic leads to compounding covariate shifts and causal confusion for learning-based models. Nevertheless, this raw baseline establishes a reference point to quantify the testing efficiency of our semantic subsets.

When the evaluation shifts to the low MprTTC scenario set, the performance of all tested policies degrades sharply. This subset filters scenarios based on inherent kinematic urgency. The collision rate for RiskIDM surges tenfold to 30.3\%, and the minimum time-to-collision drops to 1.62 seconds. QCNet experiences a nearly complete failure with a 98.3\% collision rate. This drastic performance decay validates that our kinematic risk identification pipeline successfully eliminates trivial interactions and isolates highly challenging spatiotemporal conflicts that push planning algorithms toward their dynamic limits.

The most profound algorithmic vulnerabilities are exposed when evaluating the policies against the safety of the intended functionality semantic triggers. In the visual shielding scenarios, the fundamental line-of-sight assumptions of the planning algorithms are severely disrupted. Consequently, the minimum time-to-collision plummets to critical levels across all models. The collision rates reach their peak, with the conservative RiskIDM failing 63.4\% of the time and the learning-based models exceeding 79.9\%. Furthermore, the narrow feasible area scenarios rigorously test the spatial constraint satisfaction capabilities of the algorithms. While RiskIDM struggles with high collision rates, QCNet completely fails to maintain road adherence, exhibiting a 64.0\% offroad rate and a 52.9\% traffic violation rate. These semantic edge cases force the models into complex spatial planning dilemmas that cannot be fully captured by pure kinematic thresholds.

These evaluation results unequivocally demonstrate the immense value of the SinD 2.0 dataset for autonomous driving safety validation. Relying solely on raw naturalistic data is highly inefficient for system boundary testing due to the sparsity of critical events. By leveraging our multi-level semantic annotations, researchers can directly target specific triggering conditions, thereby massively accelerating the discovery of algorithmic deficiencies and facilitating the robust development of high-reliability planning systems.

\begin{table*}[t]
\centering
\caption{Dataset-open-loop results on raw and semantic high-value scenarios.}
\label{tab:semantic_open_loop}
\begin{tabular}{ccccccc}
\hline
Scenario set                      & Policy  & MinTTC (s) & Collision (\%) & Offroad (\%) & Violation (\%) & Anomaly (\%) \\ \hline
\multirow{3}{*}{Raw naturalistic} & RiskIDM & 2.13       & 3.0            & 0.9          & 1.5            & 1.2          \\
                                  & ASAPRL  & 0.56       & 58.4           & 2.7          & 24.2           & 28.7         \\
                                  & QCNet   & 1.06       & 52.9           & 4.5          & 35.6           & 54.3         \\ \hline
\multirow{3}{*}{Low MprTTC}       & RiskIDM & 1.62       & 30.3           & 0.2          & 5.9            & 1.4          \\
                                  & ASAPRL  & 0.97       & 55.1           & 15.6         & 31.0           & 49.0         \\
                                  & QCNet   & 0.03       & 98.3           & 0.8          & 6.8            & 6.7          \\ \hline
\multirow{3}{*}{Visual shielding} & RiskIDM & 0.62       & 63.4           & 0.0          & 28.5           & 39.7         \\
                                  & ASAPRL  & 0.33       & 79.9           & 13.5         & 41.7           & 39.7         \\
                                  & QCNet   & 0.34       & 81.8           & 21.4         & 38.0           & 45.5         \\ \hline
\multirow{3}{*}{Narrow feasible}  & RiskIDM & 0.63       & 53.4           & 10.7         & 14.2           & 20.2         \\
                                  & ASAPRL  & 0.49       & 62.5           & 10.7         & 22.7           & 16.6         \\
                                  & QCNet   & 0.25       & 86.8           & 64.0         & 52.9           & 84.7         \\ \hline
\end{tabular}
\end{table*}

\subsection{Closed-loop Testing in Reactive Testing Environment}

Open-loop replay fixes the motion of non-ego agents, which induces unnatural interactive behaviors and renders this paradigm inadequate for rigorous SOTIF validation.
In Closed-loop testing, the ego can be controlled by a policy such as RiskIDM, ASAPRL, or QCNet. Nearby non-ego participants are updated by a data-driven reactive policy, allowing them to respond to ego deviations while still being anchored to the original
SinD context.
Table~\ref{tab:closed_loop} reports the results of closed-loop testing. Compared
with raw open-loop replay, all policies face a much harder test distribution.
RiskIDM remains the most stable among the evaluated controllers, with the lowest off-road rate, and lowest anomaly rate, but its collision rate
still rises to 29.6\%. ASAPRL and QCNet show higher collision and anomaly rates,
indicating that learned or prediction-driven policies can become unstable when
their actions modify the future interaction context. 
Beyond relative performance rankings, a core strength lies in diagnostic capacity of the toolchain. Identical behavior and evaluation pipelines are executable under replay and interactive closed-loop modes, enabling developers to decouple imitation inaccuracies, rule-violation faults and safety degradation induced by multi-agent interactions.

\begin{table}[t]
\centering
\caption{Closed-loop results in the Data-driven Reactive Testing Environment.}
\label{tab:closed_loop}
\scriptsize
\begin{tabular}{lrrrrrrrr}
\toprule
Policy  & MinTTC(s) & Coll.(\%) & Offroad(\%) & Viol.(\%) & Anom.(\%) \\ \midrule
RiskIDM & 1.40   & 29.6  & 0.1     & 14.5  & 10.0  \\
ASAPRL  & 0.42   & 69.1  & 1.1     & 24.4  & 33.1  \\
QCNet   & 1.11   & 50.4  & 3.7     & 23.7  & 51.2  \\ \bottomrule
\end{tabular}
\end{table}

\subsection{Implications for ADS Safety Validation}

These experiments show that SinD 2.0 is useful not only as a trajectory dataset
but also as a safety-test substrate. Offline prediction evaluates perception and
forecasting generalization under cross-domain mixed traffic. Dataset-open-loop
testing checks whether a policy can act safely when surrounded by realistic
recorded agents. Semantic open-loop testing converts rare risk triggers into
repeatable targeted stress tests. Finally, the Data-driven Reactive Testing
Environment introduces interaction feedback while preserving the original
map-signal-scene context. Together, these modes form a practical bridge from
naturalistic trajectory data to SOTIF-oriented ADS validation.

\subsection{Additional Visual Simulation Extension}

\begin{figure}[t]
    \centering
    \includegraphics[width=0.78\columnwidth]{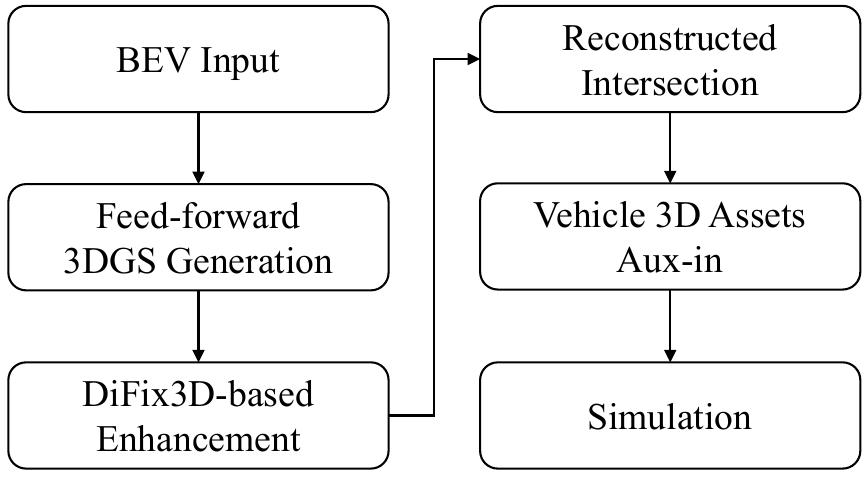}
    \caption{Auxiliary 3DGS-based visual simulation pipeline. A BEV map prior is converted into an initial 3DGS scene, refined by DiFix3D-based enhancement, combined with auxiliary vehicle 3D assets, and rendered as a photorealistic intersection environment.}
    \label{fig:3dgs_pipeline}
\end{figure}

Beyond the core BEV trajectory and closed-loop testing functions, we also
provide an auxiliary visual-simulation extension for non-BEV analysis. This
extension is intended to support future safety validation of vision-based
end-to-end autonomous driving systems, where an ego-view photorealistic
background is needed in addition to trajectory-level interaction logs. As shown
in Fig.~\ref{fig:3dgs_pipeline}, the pipeline uses the BEV map as a geometric
prior and instantiates a basic 3D scene through a feed-forward 3D Gaussian
Splatting (3DGS) generation module. A DiFix3D-based enhancement step then
refines rendering details and geometric consistency, producing a reconstructed
intersection background.

\begin{figure*}[t]
    \centering
    \includegraphics[width=0.73\textwidth]{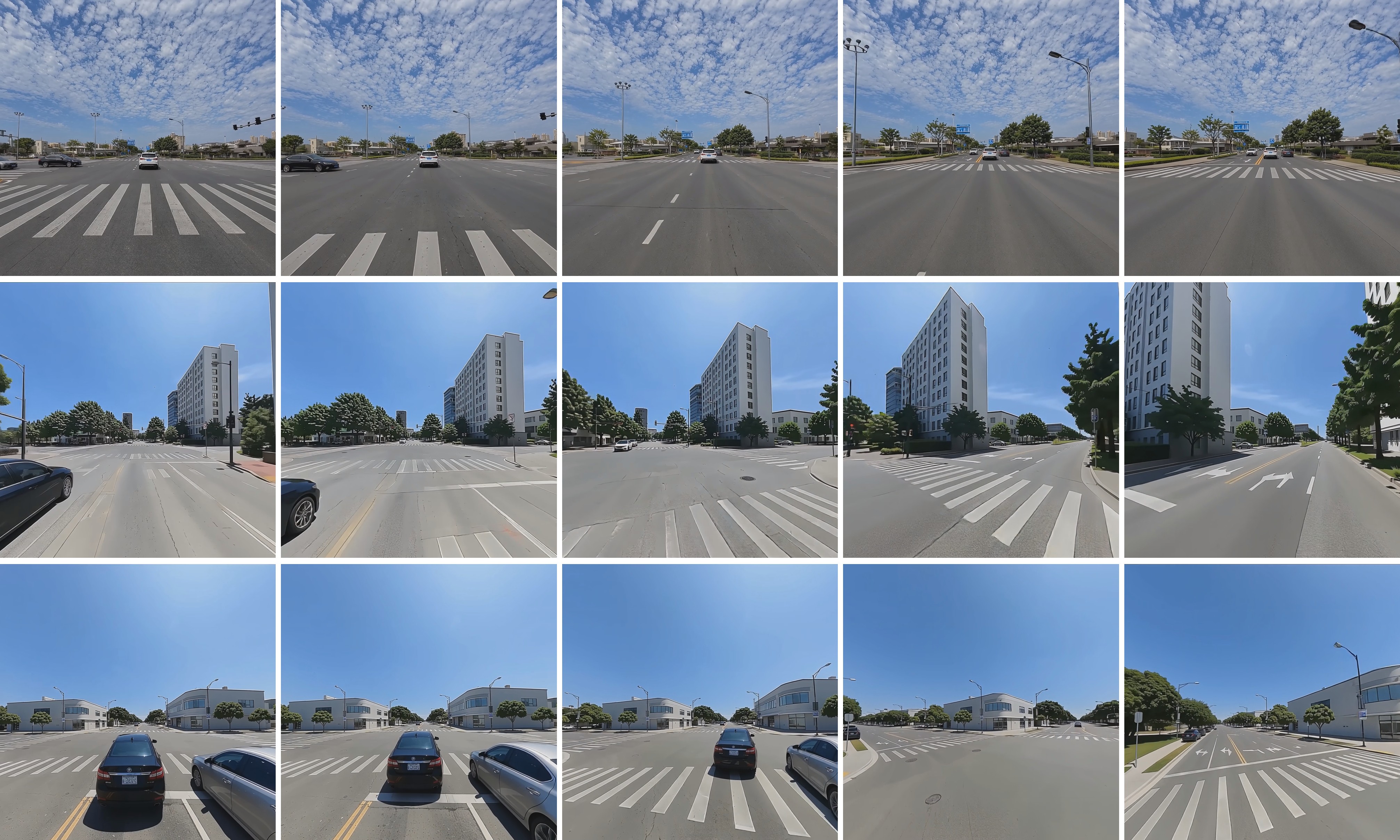}
    \caption{Egocentric visual simulation demonstrations generated by the 3DGS-based pipeline. The sequences illustrate continuous view rendering for an autonomous vehicle executing different navigational maneuvers at reconstructed intersections: proceeding straight (top row), making a right turn (middle row), and making a left turn (bottom row).}
    \label{fig:3dgs_demo}
\end{figure*}

Based on the reconstructed static background, vehicle 3D assets can be inserted
as dynamic traffic participants, enabling egocentric visual rendering under
different intersection maneuvers. Fig.~\ref{fig:3dgs_demo} illustrates example
simulation sequences for going straight, turning right, and turning left. We
treat this component as a complementary rendering capability rather than a
separate quantitative benchmark in this paper; its role is to broaden the
toolchain toward camera-view testing while the main validation results above
remain trajectory- and interaction-metric based.

\section{Conclusion}\label{sec:conclusion}
In this paper, we presented SinD 2.0, a comprehensive, multi-city drone-based traffic dataset and associated full-stack simulation toolchain specifically engineered for SOTIF-oriented autonomous driving safety validation. By capturing naturalistic multi-agent interactions across diverse urban domains and leveraging the hierarchical semantic risk annotation, we successfully bridged the gap between raw motion trajectories and explicit SOTIF triggering structures, categorizing high-risk behaviors into interpretable corner-case subsets, such as visual shielding and dynamically constrained narrow feasible areas. 

Crucially, our continuous open-loop and data-driven reactive closed-loop benchmarks yield vital technical implications for the behavior modeling community. The substantial performance degradation observed across both rule-based and learning-based planners underscores a fundamental limitation in current autonomous driving systems: standard kinematic prediction layers fail to preserve safety when confronted with spatiotemporally coupled risk stressors. Resolving these contested right-of-way bottlenecks requires a paradigm shift away from simple state-history extrapolation toward deeper semantic comprehension of visual shielding and highly interactive closed-loop evaluation. Consequently, the comprehensive toolchain provided by SinD 2.0 serves as a critical validation framework to systematically expose and evaluate these systemic policy vulnerabilities.

Despite its comprehensive framework, certain boundaries of the current methodology pave the way for future exploration. First, a subset of our structured violation annotations relies extensively on nominal map and light-binding associations, rendering the semantic extraction sensitive to upstream estimation noise. Second, while the data-driven generative controller successfully introduces reactive background traffic, it remains susceptible to behavioral approximations and semantic drifting under severe edge-case contentions. Future extensions will focus on developing hybrid evaluation architectures that intertwine neural traffic generators with rule-based safety envelopes, thereby establishing stricter behavioral guardrails for non-ego participants. Overall, we expect SinD 2.0 to catalyze collaborative efforts across academia and industry toward the verifiable deployment of highly reliable autonomous driving systems.

\appendices
\section{Detailed Criteria for Safety-Critical Event Extraction}
\label{sec:appendix_sce}

To guarantee the quality of the extracted safety-critical events defined by $\mathbb{I}_{\text{SCE}}(i, j)$, candidate trajectory pairs must satisfy a sequence of kinematic and spatiotemporal constraints.

The interacting entities are restricted to motor vehicles, cyclists, and motorcycles. We filter out stationary or near-stationary agents by requiring the total travel distance to exceed $2.0\,\text{m}$ and the maximum speed to exceed $0.5\,\text{m/s}$. For any two interacting agents, a valid candidate pair must exhibit a spatial bounding box distance of less than $20.0\,\text{m}$ and a temporal overlap within $15.0\,\text{s}$. Within these candidates, routine safe behaviors, such as basic path following, lane following, and parallel non-interacting trajectories, are programmatically excluded based on their relative spatial configurations.

The surrogate safety measures are calculated under strict boundary conditions. The PET is computed only when the distance between the closest conflict points of two trajectory segments is less than $2.5\,\text{m}$, accounting for the half-body clearance time of the entities. The robust TTC, denoted as $\text{TTC}_{\text{robust}}$, is evaluated exclusively across shared time frames. To mitigate measurement noise and transient kinematic fluctuations, the calculation requires the relative closing speed to be strictly greater than $0.5\,\text{m/s}$ and the time-to-collision values to remain valid for at least three consecutive frames. The final representative measurement is determined by extracting the $20\%$ quantile of these valid consecutive values.

% Acknowledgment is omitted in this draft.

\ifCLASSOPTIONcaptionsoff
  \newpage
\fi

\bibliographystyle{IEEEtran}
\bibliography{cas-refs}

% Author biographies are omitted in this draft.
\end{document}